\documentclass[twocolumn,10pt]{article}
\usepackage{fullpage}
\usepackage{authblk}

\usepackage{times}  %
\usepackage{helvet}  %
\usepackage{courier}  %
\usepackage[hyphens]{url}  %
\usepackage{graphicx} %
\usepackage{natbib}  %
\usepackage{caption} %
\usepackage{xcolor}         %

\usepackage[algo2e, noend, noline, ruled, linesnumbered]{algorithm2e}
\DeclareCaptionStyle{ruled}{labelfont=normalfont,labelsep=colon,strut=off} %
\usepackage{algorithm}
\usepackage{algorithmic}

\usepackage{hyperref}
\hypersetup{
    colorlinks=true,
    linkcolor=blue,
    filecolor=magenta, 
    urlcolor=cyan,
    citecolor=blue
    }
    
\usepackage{newfloat}
\usepackage{listings}
\floatstyle{ruled}
\newfloat{listing}{tb}{lst}{}
\floatname{listing}{Listing}

\usepackage{amsmath,amsfonts}

\newcommand{\actions}{A}
\newcommand{\states}{X}

\newcommand{\argmax}{\operatorname*{argmax}}

\usepackage{amsthm}

\newtheorem{theorem}{Theorem}

\theoremstyle{definition}
\newtheorem{definition}{Definition}

\newcommand{\cE}{{\mathcal E}}

\newcommand{\bbE}{{\mathbb E}}

\newcommand{\bbR}{{\mathbb R}}

\newcommand{\EE}{{\bbE}}

\newcommand{\RR}{{\bbR}}

\newcommand{\softmax}{\mathrm{softmax}} 
\newcommand{\regq}{{\bar q}}

\title{Scalable Deep Reinforcement Learning Algorithms for Mean Field Games }
\date{}

\author{%
    Mathieu Laurière$^{1,\star}$,
    Sarah Perrin$^{2,\star}$,
    Sertan Girgin$^{1,\star}$,
    Paul Muller$^3$,
    Ayush Jain$^4$,
    Theophile Cabannes$^5$,
    Georgios Piliouras$^6$,
    Julien Pérolat$^3$,
    Romuald Élie$^3$,
    Olivier Pietquin$^{1,\dagger}$,
    Matthieu Geist$^{1,\dagger}$%
  }
  \renewcommand\footnotemark{}
\date{
  $^1$Google Research, $^2$Univ. Lille, CNRS, Inria, Centrale Lille, UMR 9189 CRIStAL%
  \\
  $^3$DeepMind, $^4$ UC Berkeley, $^5$ Google research
  \\
  $^6$ Singapore University of Technology and Design
\thanks{$^\star$ equal contribution, $^\dagger$ equal contribution}
}

\begin{document}

\maketitle

\begin{abstract}

Mean Field Games (MFGs) have been introduced to efficiently approximate games with very large populations of strategic agents. Recently, the question of learning equilibria in MFGs has gained momentum, particularly using model-free reinforcement learning (RL) methods. One limiting factor to further scale up using RL is that existing algorithms to solve MFGs require the mixing of approximated quantities such as strategies or $q$-values. This is far from being trivial in the case of non-linear function approximation that enjoy good generalization properties, \textit{e.g.} neural networks. We propose two methods to address this shortcoming. The first one learns a mixed strategy from distillation of historical data into a neural network and is applied to the Fictitious Play algorithm. The second one is an online mixing method based on regularization that does not require memorizing historical data or previous estimates. It is used to extend Online Mirror Descent. We demonstrate numerically that these methods efficiently enable the use of Deep RL algorithms to solve various MFGs. In addition, we show that these methods outperform SotA baselines from the literature.

\end{abstract}

\everypar{\looseness=-1}

\vspace{-0.25\baselineskip}
\section{Introduction}

Despite outstanding success of machine learning in numerous applications~\citep{goodfellow2016deep,goodfellow2020generative}, learning in games remains difficult because two or more agents require the consideration of non-stationary environments. In particular, it is hard to scale in terms of number of agents because one has to keep track of all agents' behaviors, leading to a prohibitive combinatorial complexity \citep{daskalakis2008, Tuyls_Weiss_2012}. 
Recently, Mean Field Games (MFGs) have brought a new perspective by replacing the atomic agents by their distribution. Introduced concurrently by \citet{MFG_lasry-lions} and \citet{2006huang-mfg}, the mean field approximation relies on symmetry and homogeneity assumptions, and enables to scale to an infinite number of agents. Solving MFGs is usually done by solving a forward-backward system of partial (or stochastic) differential equations, which represent the dynamics of both the state distribution and the value function. However, traditional numerical methods (see  \textit{e.g.}~\citep{MR2679575,MR3148086,MR3772008} and ~\citep{achdou2020meanfieldgamesnumericalcetraro} for a survey) do not scale well in terms of the complexity of the state and action spaces. Recently, several deep learning methods for MFGs have been developed to tackle high-dimensional problems (see \textit{e.g.}~\citep{fouque2020deep,carmona2019convergencefinite,germain2019numerical,cao2020connecting} and~\citep{carmonalauriere2021deepmfgsurvey} for a survey). Nevertheless, these methods require the model (dynamics and rewards) to be completely known.

On the other hand, Reinforcement Learning (RL) has proved very efficient to solve Markov Decision Processes (MDPs) and games with complex structures,  from \textit{e.g.} chess~\citet{campbell2002deep} to multi-agent systems~\citep{lanctot2017unified}. To tackle problems with highly complex environments, RL algorithms can be combined with efficient function approximation. Deep neural networks with suitable architectures are arguably the most popular choice, thanks to their ease of use and their generalization capabilities. Deep RL has brought several recent breakthrough results, \textit{e.g.}, Go~\citep{silver2016mastering}, atari~\citep{mnih2013playingatari},  poker~\citep{brown2018superhuman,moravvcik2017deepstack} or even video games such as Starcraft~\citep{vinyals2019grandmaster}.

Many recent works have combined MFGs with RL to leverage their mutual potential -- albeit mostly without deep neural nets thus far. 
The intertwinement of MFGs and RL happens through an \emph{optimization} (or learning) procedure. The simplest algorithm of this type is the (Banach-Picard) fixed-point approach, consisting in alternating a best response computation against a given population distribution with an update of this distribution~\citep{2006huang-mfg}. However, this method fails in many cases by lack of contractivity as proved by~\citet{pmlr-cui21-approximately}. Several other procedures have thus been introduced, often inspired by game theory or optimization algorithms. Fictitious Play (FP) and its variants average either distributions or policies (or both) to stabilize convergence \citep{hadikhanloo_fictitious-play,elie2020convergence,perrin2020fictitious,pmlr-xie21-learning,perrin2021generalizationmfg,perrin2021mfgflockrl}, whereas Online Mirror Descent (OMD) \citep{hadikhanloo2017learningnonatomic,perolat2021scaling} relies on policy evaluation. Other works have leveraged regularization \citep{anahtarci2020qregmfg,pmlr-cui21-approximately,guoxuzariphopoulou2020entropyregmfg} to ensure convergence, at the cost of biasing the Nash equilibrium.  
These methods require to sum or average some key quantities: FP needs to average the distributions, while OMD needs to sum $Q$-functions. This is a key component of most (if not all) smoothing methods and needs to be tackled efficiently. These operations are simple when the state space is finite and small, and the underlying objects can be represented with tables or linear functions. However, there is no easy and efficient way to sum non-linear approximations such as neural networks, which raises a major challenge when trying to combine learning methods (such as FP or OMD) with deep RL. 

The main contribution of the paper is to solve this important question in dynamic MFGs. We propose two algorithms. The first one, that we name Deep Average-network Fictitious Play (D-AFP), builds on FP and uses the Neural Fictitious Self Play (NFSP) approach~\citep{heinrichsilver2016deepnfsp} to compute a neural network approximating an average over past policies. The second one is Deep Munchausen Online Mirror Descent (D-MOMD), inspired by the Munchausen reparameterization of \citet{vieillard_munchausen_neurips_2020}.
We prove that in the exact case, Munchausen OMD is equivalent to OMD.
Finally, we conduct numerical experiments and compare D-AFP and D-MOMD with SotA baselines adapted to dynamic MFGs. 
We find that D-MOMD converges faster than D-AFP on all tested games from the literature, which is consistent with the results obtained for exact algorithms (without RL) in \citep{perolat2021scaling,geist2021curl}.

\vspace{-0.25\baselineskip}
\section{Background}

\vspace{-0.25\baselineskip}
\subsection{Mean Field Games}

A Mean Field Game (MFG) is a strategic decision making problem with a continuum of identical and anonymous players. In MFGs, it is possible to select a \emph{representative player} and to focus on its policy instead of considering all players individually, which simplifies tremendously the computation of an equilibrium. We place ourselves in the context of a game and are interested in computing Nash equilibria. We stress that we consider a \emph{finite horizon} setting as it encompasses a broader class of games, which needs time-dependant policies and distributions. We denote by $N_T$ the finite time horizon and by $n$ a generic time step. We focus on finite state space and finite action space, denoted respectively by $\states$ and $\actions$.  
We denote by $\Delta_\states$ the set of probability distributions on $\states$. Each distribution can be viewed as a vector of length $|\states|$. Then, the game is characterized by one-step reward functions $r_n: \states \times \actions \times \Delta_\states \to \RR$, and transition probability functions $p_n: \states \times \actions \times \Delta_\states \to \Delta_\states$, for $n=0,1,\dots,N_T$. The third argument corresponds to the current distribution of the population. An initial distribution $m_0$ of the population is set and will remain fixed over the paper. 
The two main quantities of interest are the policy of the representative player $\pi = (\pi_n)_n\in(\Delta_{\actions}^\states)^{N_T+1}$ and the distribution flow (\textit{i.e.} sequence) of agents $\mu = (\mu_n)_n\in\Delta_{\states}^{N_T+1}$. Let us stress that we denote the time of the game $n$ and that $\pi$ and $\mu$ are thus time-dependant objects.
Given a population mean field flow $\mu$, the goal for a representative agent is to maximize over $\pi$ the total reward:
\begin{align*}
    &J(\pi, \mu) = \mathbb{E}_\pi \Big[\sum_{n=0}^{N_T} r_n(x_n, a_n, \mu_n) \Big| x_0 \sim m_0\Big]
\\    &\hbox{s.t.: } a_n \sim \pi_n(\cdot|x_n), \, x_{n+1} \sim p_n(\cdot|x_n,a_n,\mu_n), \, n \ge 0.
\end{align*}
Note that the reward $r_n$ together with the transition function $p_n$ are functions of the population distribution at the current time step $\mu_n$, which encompasses the mean field interactions. 
A policy $\pi$ is called a best response (BR) against a mean field flow $\mu$ if it is a maximizer of $J(\cdot,\mu)$. We denote by $BR(\mu)$ the set of best responses to the mean field flow $\mu$.

Given a policy $\pi$, a mean field flow $\mu$ is said to be induced by $\pi$ if: $\mu_0 = m_0$ and for $n=0,\dots,N_T-1$,
\begin{equation*}
    \mu_{n+1}(x) = \sum_{x',a'}\mu_n(x') \pi_n(a'|x') p_n(x|x',a',\mu_n),
\end{equation*}
which we can simply write $\mu_{n+1} = P^{\mu_n,\pi_n}_n \mu_n$, where $P^{\mu_n,\pi_n}_n$ is the transition matrix of $x_n$. We denote by $\mu^\pi$ or $\Phi(\pi) \in \Delta_\states^{N_T+1}$ the mean-field flow induced by $\pi$. 

\begin{definition}
A pair $(\hat\pi,\hat\mu)$ is a (finite horizon) Mean Field Nash Equilibrium (MFNE) if (1) $\hat\pi$ is a BR against $\hat\mu$, and (2) $\hat\mu$ is induced by $\hat\pi$.
\end{definition}
Equivalently, $\hat\pi$ is a fixed point of the map $BR \circ \Phi$. Given a mean field flow $\mu$, a representative player faces a traditional MDP, which can be studied using classical tools. The value of a policy can be characterized through the $Q$-function defined as: $Q^{\pi, \mu}_{N_T+1}(x,a) = 0$ and for $n \le N_T$,
\begin{equation*}
    Q^{\pi, \mu}_{n}(x,a) = \EE\Big[\sum_{n' \ge n} r_{n'}(x_{n'}, a_{n'}, \mu_{n'}) \Big|  (x_n, a_n)=(x, a) \Big].
\end{equation*}
It satisfies the Bellman equation:  for $n \le  N_T$, 
 \begin{equation}
     \label{eq:eval-Bellman-Q}
     Q^{\pi, \mu}_{n}(x,a) = r_{n}(x, a, \mu_{n}) + \EE_{x',a'}[Q^{\pi, \mu}_{n+1}(x',a')], 
 \end{equation} 
 \looseness=-1
 where $x' \sim p(\cdot|x,a,\mu_n)$ and $a' \sim \pi_n(\cdot|x,a,\mu_n)$,   with the convention $Q^{\pi, \mu}_{N_T+1}(\cdot,\cdot) = 0$. 
The optimal $Q$-function $Q^{*,\mu}$ is the value function of any best response $\pi^*$ against $\mu$. It is defined as $Q^{*,\mu}_n(x,a) = \max_{\pi} Q^{\pi,\mu}_n(x,a)$ for every $n,x,a$, and it satisfies the optimal Bellman equation: for $n \le  N_T$, 
\begin{equation}
     \label{eq:opt-Bellman-Q}
     Q^{*, \mu}_{n}(x,a) = r_{n}(x, a, \mu_{n}) + \EE_{x',a'}[\max_{a'}Q^{*, \mu}_{n+1}(x',a')],  
 \end{equation} 
 where $Q^{*, \mu}_{N_T+1}(\cdot,\cdot) = 0$.

\subsection{Fictitious Play}
\label{sec:fictitious-play}
The most straightforward method to compute a MFNE is to iteratively update in turn the policy $\pi$ and the distribution $\mu$, by respectively computing a BR and the induced mean field flow. The BR can be computed with the backward induction of~\eqref{eq:opt-Bellman-Q}, if the model is completely known.  We refer to this method as \textit{Banach-Picard (BP) fixed point iterations}. See Alg.~\ref{alg:fixed-point} in appendix for completeness. The convergence is ensured as soon as the composition $BR \circ \Phi$ is a strict contraction~\citep{2006huang-mfg}.  
However, this condition holds only for a restricted class of games and, beyond that, simple fixed point iterations typically fail to converge and oscillations appear~\citep{pmlr-cui21-approximately}.  

To address this issue, a memory of past plays can be added. The \textit{Fictitious Play (FP) algorithm}, introduced by \citet{brown1951iterative} computes the new distribution at each iteration by taking the average over all past distributions instead of the latest one. This stabilizes the learning process so that convergence can be proved for a broader class of games under suitable assumptions on the structure of the game such as potential structure~\citep{hadikhanloo_fictitious-play,geist2021curl} or monotonicity~\citep{perrin2020fictitious}. The method can be summarized as follows: after initializing $Q^0_n$ and $\pi_n^0$ for $n=0,\dots,N_T$, repeat at each iteration $k$: 
    \begin{equation*}
        \left\{
        \begin{array}{llll}
        \mbox{1. Distribution update: } \mu^k = \mu^{\pi^k}, \bar\mu^{k} = \frac{1}{k-1}\sum_{i=1}^{k-1}\mu^{i} \\
        \mbox{2. $Q$-function update: } Q^k = Q^{*, \bar\mu^k}\\
        \mbox{3. Policy update: } \pi^{k+1}_n(.|x) = \argmax_{a} Q^k_n(x,a).
        \end{array}
        \right.
    \end{equation*}
In the distribution update, $\bar\mu^{k}$ corresponds to the population mean field flow obtained if, for each $i=1,\dots,k-1$, a fraction $1/{(k-1)}$ of the population follows the policy $\pi^i$ obtained as a BR at iteration $i$. At the end, the algorithm returns the latest mean field flow $\mu^k$ as well as a policy that generates this mean field flow. This can be achieved either through a single policy or by returning the vector of all past BR, $(\pi^i)_{i=1,\dots,k-1}$, from which $\bar\mu^{k}$ can be recovered.  
See Alg.~\ref{alg:Fictitious-play} in appendix for completeness.

\subsection{Online Mirror Descent}
\label{sec:background-OMD}

The aforementioned methods are based on computing a BR at each iteration. Alternatively, we can follow a policy iteration based approach and simply \emph{evaluate} a policy at each iteration. In finite horizon, this operation is less computationally expensive than computing a BR because it avoids a loop over the actions to find the optimal one.  

The \textit{Policy Iteration (PI) algorithm} for MFG~\citep{cacace2020policy} consists in repeating, from an initial guess $\pi^0,\mu^0$, the update: at iteration $k$, first evaluate the current policy $\pi^k$ by computing $Q^{k+1} = Q^{\pi^k,\mu^k}$, then let $\pi^{k+1}$ be the greedy policy such that  $\pi^{k+1}(\cdot|x)$ is a maximizer of $Q^{k}(x, \cdot)$.  The evaluation step can be done with the backward induction~\eqref{eq:eval-Bellman-Q}, provided the model is known.  See Alg.~\ref{alg:policy-iteration} in appendix for completeness.

Here again, to stabilize the learning process, one can rely on information from past iterations. Using a weighted sum over past $Q$-functions yields the so-called \textit{Online Mirror Descent (OMD) algorithm} for MFG, which can be summarized as follows: after initializing $q^0_n$ and $\pi_n^0$ for $n=0,\dots,N_T$, repeat at each iteration $k$: 
    \begin{equation*}
        \left\{
        \begin{array}{llll}
        \mbox{1. Distribution update: } \mu^k = \mu^{\pi^k} \\
        \mbox{2. $Q$-function update: } Q^k = Q^{\pi^k, \mu^k}\\
        \mbox{3. Regularized $Q$-function update: } \regq^{k+1} = \regq^k + \frac{1}{\tau} Q^{k}\\
        \mbox{4. Policy update: } \pi^{k+1}_n(\cdot|x) = \softmax(\regq_n^{k+1}(x,\cdot)).
        \end{array}
        \right.
    \end{equation*}
For more details, see Alg.~\ref{Algo:OMD-exact} in appendix. Although we focus on softmax policies in the sequel, other conjugate functions of steep regularizers could be used in OMD, see \citet{perolat2021scaling}. 
 
This method is known to be empirically faster than FP, as illustrated by~\citet{perolat2021scaling}. Intuitively, this can be explained by the fact that the learning rate in FP is of the order $1/k$ so this algorithm is slower and slower as the number of iterations increases.

\subsection{Deep Reinforcement Learning}

\looseness=-1
Reinforcement learning aims to solve optimal control problems when the agent does not know the model (\textit{i.e.}, $p$ and $r$) and must learn through trial and error by interacting with an environment. In a finite horizon MFG setting, we assume that a representative agent is encoded by a policy $\pi$, either explicitly or implicitly (through a $Q$-function) and can realize an episode, in the following sense: for $n=0,\dots,N_T$, the agent observes $x_n$ (with $x_0 \sim m_0$), chooses action $a_n \sim \pi_n(\cdot|x_n)$, and the environment returns a realization of $x_{n+1} \sim p_n(\cdot|x_n,a_n,\mu_n)$ and $r_n(x_n,a_n,\mu_n)$. Note that the agent does not need to observe directly the mean field flow $\mu_n$, which simply enters as a parameter of the transition and cost functions $p_n$ and $r_n$. 

\looseness=-1
Based on such samples, the agent can approximately compute the $Q$-functions $Q^{\pi,\mu}$ and $Q^{*,\mu}$ following~\eqref{eq:eval-Bellman-Q} and~\eqref{eq:opt-Bellman-Q} respectively where the expectation is replaced by Monte-Carlo samples. In practice, we often use trajectories starting from time $0$ and state $x_0 \sim m_0$ instead of starting from any pair $(x,a)$. Vanilla RL considers infinite horizon, discounted problems and looks for a stationary policy, whereas we consider a finite-horizon setting with non-stationary policies. To simplify the implementation, we treat time as part of the state by considering  $(n,x_n)$ as the state. We can then use standard $Q$-learning. However, it is important to keep in mind that the Bellman equations are not fixed-point equation for some stationary Bellman operators.

When the state space is large, it becomes impossible to evaluate precisely every pair $(x,a)$. Motivated by both memory efficiency and generalization, we can approximate the $Q$-functions by non linear functions such as neural networks, say $Q^{\pi,\mu}_\theta$ and $Q^{*,\mu}_\theta$, parameterized by $\theta$. Then, the quantities in~\eqref{eq:eval-Bellman-Q} and~\eqref{eq:opt-Bellman-Q} are replaced by the minimization of a loss to train the neural network parameters $\theta$. Namely, treating time as an input, one minimizes over $\theta$ the quantities 
\begin{align*}
    &\hat{\EE}\left[ \left| Q^{\pi, \mu}_{\theta,n}(x,a) - r_{n}(x, a, \mu_{n}) - Q^{\pi, \mu}_{\theta_t,n+1}(x',a') \right|^2 \right]
    \\
    &\hat{\EE}\left[ \left| Q^{*, \mu}_{\theta,n}(x,a) - r_{n}(x, a, \mu_{n}) - \max_{a'}Q^{*, \mu}_{\theta_t,n+1}(x',a') \right|^2 \right],
\end{align*}
where $\hat{\EE}$ is an empirical expectation based on Monte Carlo samples and $\theta_t$ is the parameter of a target network.

\section{Deep Reinforcement Learning for MFGs}

To develop scalable methods for solving MFGs, a natural idea consists in combining the above optimization methods (FP and OMD) with deep RL. This requires summing or averaging policies or distributions, and  induces hereby a major challenge as they are approximated by non linear operators, such as neural networks. In this section, we develop innovative and scalable solutions to cope with this. In the sequel, we denote $Q_\theta((n,x),a)$ with the time in the input when we refer to the neural network $Q$-function.

\subsection{Deep Average-network Fictitious Play}

To develop a model-free version of FP, one first needs to compute a BR at each iteration, which can be done using standard deep RL methods, such as DQN \citep{mnih2013playingatari}. A policy that generates the average distribution over past iterations can be obtained by simply keeping in memory all the BRs from past iterations. This approach has already been used successfully \textit{e.g.} by \citet{perrin2020fictitious,perrin2021mfgflockrl}. However, it requires a memory that is linear in the number of iterations and each element is potentially large (\textit{e.g.}, a deep neural network), which does not scale well. Indeed, as the complexity of the environment grows, we expect FP to need more and more iterations to converge, and hence the need to keep track of a larger and larger number of policies.

An alternative approach is to learn along the way the policy generating the average distribution. We propose to do so by keeping a buffer of state-action pairs generated by past BRs and learning the average policy by minimizing a categorical loss. To tackle potentially complex environments, we rely on a neural network representation of the policy. This approach is inspired by the Neural Fictitious Self Play (NFSP) method~\citep{heinrichsilver2016deepnfsp}, developed initially for imperfect information games with a finite number of players, and adapted here to the MFG setting. The proposed algorithm, that we call D-AFP because it learns an average policy, is summarized in Alg.~\ref{algo:deep-FP-avgBR}.  Details are in Appx.~\ref{sec:deep-rl-algos}.

\begin{algorithm}[tb]
   \caption{D-AFP } 
   \label{algo:deep-FP-avgBR}
\begin{algorithmic}[1] 
   \STATE Initialize an empty reservoir buffer $\mathcal{M}_{SL}$ for supervised learning of average policy
   \STATE Initialize the parameters $\bar\theta^{0}$
   \FOR{$k=1,\dots,K$}
      \STATE \textbf{1. Distribution:} Generate $\bar\mu^k$ with 
      $\bar\pi_{\bar\theta^{k-1}}$  
      \STATE \textbf{2. BR:} Train $\hat\pi_{\theta^{k}}$ against  
      $\bar\mu^{k-1}$, \textit{e.g.} using DQN
      \STATE Collect $N_{samples}$ state-action using $\hat\pi_{\theta^{k}}$ and add them to $\mathcal{M}_{SL}$
      \STATE \textbf{3. Average policy:} Update $\bar\pi_{\bar\theta^{k}}$ by adjusting $\bar\theta^k$ (through gradient descent) to minimize: 
    $$
      \quad\quad  \mathcal{L}(\bar\theta) = \EE_{(s,a) \sim \mathcal{M}_{SL}}\left[-\log \left( \bar\pi_{\bar\theta}(a|s)\right)\right], 
    $$
      where $\bar\pi_{\bar\theta}$ is the neural net policy with parameters $\bar\theta$ 
   \ENDFOR
   \STATE Return $\bar\mu^{K}, \bar\pi_{\bar\theta^{K}}$
\end{algorithmic}
\end{algorithm}

\looseness=-1
This allows us to learn an approximation of the MFNE policy with a single neural network instead of having it indirectly through a collection of neural networks for past BRs. After training, we can use this neural average policy in a straightforward way. Although the buffer is not needed after training, a drawback of this method is that during the training it requires to keep a buffer whose size increases linearly with the number of iterations. This motivates us to investigate a modification of OMD which is not only empirically faster, but also less memory consuming.

\subsection{Deep Munchausen Online Mirror Descent}

We now turn our attention to the combination of OMD and deep RL. One could simply use RL for the policy evaluation step by estimating the $Q$-function using equation~\eqref{eq:eval-Bellman-Q}. However, it is not straightforward to train a neural network to approximate the cumulative $Q$-function. To that end, we propose a reparameterization allowing us to compute the cumulative $Q$-function in an implicit way, building on the Munchausen trick from \citet{vieillard_munchausen_neurips_2020} for classical RL (with a single agent and no mean-field interactions).

\paragraph{Reparameterization in the exact case.}

OMD requires summing up $Q$-functions to compute the regularized $Q$-function $\regq$. 
However, this quantity $\regq$ is hard to approximate as there exists no straightforward way to sum up neural networks.  
We note that this summation is done by \citet{perolat2021poincare} via the use of the NeuRD loss. However, this approach relies on two types of approximations, as one must learn the correct $Q$-function, but also the correct sum. This is why we instead transform the OMD formulation into Munchausen OMD, which only relies on one type of approximation. We start by describing this reparameterization in the exact case, \textit{i.e.}, without function approximation. We show that, in this case, the two formulations are equivalent.

We consider the following modified Bellman equation:
\begin{equation}
\label{eq:Munchausen-Q-bellman}
\begin{cases}
    &\tilde Q_{N_T+1}^{k+1}(x,a) = 0
    \\
    &\tilde Q_{n-1}^{k+1}(x,a) = r(x,a,\mu^k_{n-1}) \,{\color{red}+ \tau\ln\pi_{n-1}^k(a|x) } 
    \\
    &\quad + \EE_{x',a'}\Big[\tilde Q_n^{k+1}(x',a')  \,{\color{blue}- \tau \ln\pi^k_n(a'|x')} \Big],
\end{cases}
\end{equation}
where $x' \sim p_n(\cdot|x,a,\mu_{n-1}^k)$ and $a'\sim\pi_n^k(\cdot|x')$.
The {\color{red} red term} penalizes the policy for deviating from the one in the previous iteration, $\pi_{n-1}^k$, while the {\color{blue} blue term} compensates for this change in the backward induction, as we will explain in the proof of Thm~\ref{thm:equivalence-OMD-Munchausen} below.

The Munchausen OMD (MOMD) algorithm for MFG is as follows: after initializing $\pi^0$, repeat for $k \geq 0$:
    \begin{equation*}
        \left\{
        \begin{array}{lll}
        \mbox{Distribution update: } \mu^k = \mu^{\tilde\pi^k}\\
        \mbox{Regularized $Q$-function update: } \tilde Q_n^{k+1} \mbox{ as in~\eqref{eq:Munchausen-Q-bellman}}\\
        \mbox{Policy update: } \tilde\pi^{k+1}_n(\cdot|x) = \softmax(\frac{1}{\tau} \tilde Q_n^{k+1}(x,\cdot)).
        \end{array}
        \right.
    \end{equation*}

\begin{theorem}
\label{thm:equivalence-OMD-Munchausen}
MOMD is equivalent to OMD in the sense that $\tilde\pi^k = \pi^k$ for every $k$.
\end{theorem}
As a consequence, despite seemingly artificial log terms, this method does not bias the Nash equilibrium (in contrast with, \textit{e.g.}, \citet{pmlr-cui21-approximately} and \citet{pmlr-xie21-learning}). Thanks to this result, MOMD enjoys the same convergence guarantees as OMD, see~\citep{hadikhanloo2017learningnonatomic,perolat2021scaling}.

\begin{proof}
\textbf{Step 1: Softmax transform. } 
We first replace this projection by an equivalent viewpoint based on the Kullback-Leibler (KL) divergence, denoted by $KL(\cdot \| \cdot)$. We will write $Q_n^{k+1} = Q_n^{\pi^k, \mu^{\pi^k}}$ for short and $Q_n^0 = \regq_n^0$. 
We have: $\regq_n^{k+1} = \frac{1}{\tau} \sum_{\ell=0}^{k+1} Q_n^\ell$.
We take $\pi_n^0$  
as the uniform policy over actions, in order to have a precise equivalence with the following for $\regq_n^0 = 0$.  We could consider any $\pi_n^0$ with full support, up to a change of initialization $\regq_n^0$. We have:
\begin{align}
\pi_n^{k+1}(\cdot|x) & = \text{softmax}\left(\frac{1}{\tau} \sum_{\ell=0}^{k+1} Q_n^\ell(\cdot|x)\right)
    \notag
    \\
    & = \argmax_{\pi \in \Delta_A}\Big(\langle\pi, Q_n^{k+1}(x,\cdot)\rangle - \tau KL(\pi \| \pi_n^k(\cdot|x))\Big),
    \label{eq:omd-KL-reformulation}
\end{align}
where $\langle\cdot,\cdot\rangle$ denotes the dot product.

Indeed, this can be checked by induction, using the Legendre-Fenchel transform: omitting $n$ and $x$ for brevity,
\begin{equation*}
    \begin{split}
        \pi^{k+1} & \propto \pi^k e^{\frac{1}{\tau} q^{k+1}} \\
        & \propto \pi^{k-1} e^{\frac{1}{\tau} Q^{k}} e^{\frac{1}{\tau} Q^{k+1}} = \pi^{k-1} e^{\frac{1}{\tau} (Q^{k}+Q^{k+1})}\\
        & \propto ...  \propto e^{\regq^{k+1}}.
    \end{split}
\end{equation*}

\textbf{Step 2: Munchausen trick.} Simplifying a bit notations,
\begin{equation*}
    \begin{split}
        \pi^{k+1} & = \argmax(\langle \pi, Q^{k+1}\rangle - \tau KL(\pi \| \pi^k)) \\ 
        & = \argmax(\langle \pi, \underbrace{Q^{k+1} + \tau \ln \pi^k}_{\tilde Q^{k+1}}\rangle \underbrace{- \tau \langle \pi, \ln \pi}_{+ \tau \mathcal{H}(\pi)} \rangle) \\ 
        & = \text{softmax}\Big(\frac{1}{\tau} \tilde Q^{k+1}\Big)
    \end{split}
\end{equation*}
where $\mathcal{H}$ denotes the entropy and we defined $\tilde Q_n^{k+1} = Q_n^{k+1} + \tau \ln \pi^k_n$. Since $Q_n^{k+1}$ satisfies the Bellman equation~\eqref{eq:eval-Bellman-Q} with $\pi$ replaced by $\pi^k$ and $\mu$ replaced by $\mu^k$, we deduce that $\tilde Q_n^{k+1}$ satisfies~\eqref{eq:Munchausen-Q-bellman}.
\end{proof}

\textit{Remark:} In OMD, $\frac{1}{\tau}$ (denoted $\alpha$ in the original paper~\citep{perolat2021scaling}), is homogeneous to a learning rate. In the MOMD formulation, $\tau$ can be seen as a temperature.

\paragraph{Stabilizing trick. }

We have shown that MOMD is equivalent to OMD. However, the above version of Munchausen sometimes exhibits numerical instabilities. This is because, if an action $a$ is suboptimal in a state $x$, then $\pi_n^k(a|x) \to 0$ as $k\to+\infty$, so $\tilde Q^k(x,a)$ diverges to $-\infty$ due to the relation $\tilde Q_n^{k+1} = Q_n^{k+1} + \tau \ln\pi_n^k$. This causes issues even in the tabular setting when we get close to a Nash equilibrium, due to numerical errors on very large numbers. 
To avoid this issue, we introduce another parameter, denoted by $\alpha \in [0,1]$ and we consider the following modified Munchausen equation:
\begin{align}
    &\check Q^{k+1}_{n-1}(x,a) = r_{n-1}(x, a, \mu^{k}_{n-1}) \,{\color{red}+ \alpha\tau\log(\pi^{k-1}_{n-1}(a|x))}\notag 
            \\
    &\quad + \EE_{x',a'}\Big[\tilde Q_n^{k+1}(x',a')  \,{\color{blue}- \tau \ln\pi^k_n(a'|x')} \Big],
\label{eq:Munchausen-alpha}
\end{align}
where $x' \sim p_n(\cdot|x,a,\mu_{n-1}^k)$ and $a'\sim\pi_n^k(\cdot|x')$.
In fact, such iterations have a natural interpretation as they can be obtained by applying OMD to a regularized problem in which, when using policy $\pi$, a penalty $-(1-\alpha)\tau\log(\pi_n(\cdot|x_n))$ is added to the reward $r_n(x_n,a_n,\mu_n)$. Details are provided in Appx.~\ref{sec:details-Munchausen-alpha}.

\paragraph{Deep RL version. }

Motivated by problems with large spaces, we then replace the Munchausen $Q$-function at iteration $k$, namely $\check Q^k$, by a neural network whose parameters $\theta^k$ are trained to minimize a loss function representing~\eqref{eq:Munchausen-alpha}. Since we want to learn a function of time, we consider $(n,x)$ to be the state. To be specific, given samples of transitions 
\begin{equation*}
    \left\{\Big((n_i,x_i), a_i, r_{n_i}(x_i, a_i, \mu^{k}_{n_i}), (n_i+1,x'_i)\Big)\right\}_{i=1}^{N_B},
\end{equation*}
with $x_i' \sim p_{n_i}(x_i, a_i, \mu^{k}_{n_i})$, 
the parameter $\theta^k$ is trained using stochastic gradient descent to minimize the empirical loss:
\begin{equation*}
    \begin{split}
     &\frac{1}{N_B}\sum_{i}\Big|\check Q_{\theta}((n_i,x_i),a_i) - T_i\Big|^2,
    \end{split}
\end{equation*}
where the target $T_i$ is:
\begin{align} 
    T_i &= - r_{n_i}(x_i, a_i, \mu^{k}_{n_i}) 
      \,{\color{red} - \alpha\tau\log(\pi^{k-1}(a_i|(n_i,x_i)))} 
      \notag
            \\
            &\quad - \sum_{a'} \pi^{k-1}(a'|(n_i+1,x'_i)) \Big[ \check Q_{\theta^{k-1}}((n_i+1,x'_i),a')
            \notag
            \\
            &\quad \,{\color{blue}- \tau\log(\pi^{k-1}(a'|(n_i+1,x'_i)))} \Big]. 
\label{eq:Munch-target}
\end{align}
Here the time $n$ is passed as an input to the $Q$-network along with $x$, hence our change of notation. This way of learning the Munchausen $Q$-function is similar to DQN, except for two changes in the target: (1) it incorporates the penalization for deviating from the previous policy, and (2) we do not take the argmax over the next action but an average according to the previous policy. A simplified version of the algorithm is presented in Alg.~\ref{Algo:M-OMD-exact} and more details are provided in Appx.~\ref{sec:deep-rl-algos}.

\begin{algorithm}[h!]
   \caption{D-MOMD}
   \label{Algo:M-OMD-exact}
\begin{algorithmic}[1]
   \STATE {\bfseries Input:} Munchausen parameters $\tau$ and $\alpha$;  numbers of OMD iterations $K$ and DQN estimation iterations $L$
   \STATE {\bfseries Output:} cumulated $Q$ value function, policy $\pi$
   \STATE Initialize the parameters $\theta^{0}$ 
   \STATE Set $\pi^{0}(a|(n,x)) 
   = \softmax\Big(\frac{1}{\tau} \check Q_{\theta^{0}}((n,x), \cdot)\Big)(a)$  
   \FOR{$k=1,\dots,K$}
   \STATE \textbf{1. Distribution:} Generate $\mu^{k}$ with $\pi^{k-1}$  
   \STATE \textbf{2. Value function:} 
   Initialize $\theta^{k}$ 
   \FOR{$\ell=1,\dots,L$}
        \STATE Sample a minibatch of $N_B$ transitions: 
            $\left\{\Big((n_i,x_i), a_i, r_{n_i}(x_i, a_i, \mu^{k}_{n_i}), (n_i+1,x'_i)\Big)\right\}_{i=1}^{N_B}
            $ 
         with $n_i \leq N_T$, $x'_i \sim p_{n_i}(\cdot | x_i, a_i, \mu^k_{n_i})$ and $a_i$ is chosen by an $\epsilon-$greedy policy based on $\check Q_{\theta^{k}}$ 
         
        \STATE Update $\theta^{k}$ with one gradient step of: 
        
        $
            \quad\quad\theta \mapsto \frac{1}{N_B} \sum_{i=1}^{N_B} \Big|\check Q_{\theta}((n_i,x_i), a_i) - T_{i}\Big|^2
            $
            
        where $T_{i}$ is defined in~\eqref{eq:Munch-target}  
        \ENDFOR
      \STATE {\bf 3. Policy:} for all $n, x, a$, let 

    $
   \quad\quad\pi^{k}(a|(n,x)) = \softmax\Big(\frac 1 \tau \check Q_{\theta^{k}}((n,x), \cdot)\Big)(a) 
  $ 
    \ENDFOR
   \STATE Return $\check Q_{\theta^{K}}, \pi^{K}$
\end{algorithmic}
\end{algorithm}

\section{Experiments}

In this section, we first discuss the metric used to assess quality of learning, detail baselines to which we compare our algorithms, and finally present numerical results on diverse and numerous environments. The code for Deep Munchausen OMD is available in OpenSpiel~\citep{lanctot2019openspiel}.\footnote{See \url{https://github.com/deepmind/open_spiel/blob/master/open_spiel/python/mfg/algorithms/munchausen_deep_mirror_descent.py}.}
 
\subsection{Exploitability } 
To assess the quality of a learnt equilibrium,
we check whether, in response to the reward generated by the population MF flow, a typical player can improve their reward by deviating from the policy used by the rest of the population. This is formalized through the notion of exploitability.

The exploitability of a policy $\pi$ is defined as:
\begin{equation*} 
    \cE(\pi) = \max \limits_{\pi'} J(\pi'; \mu^{\pi}) - J(\pi; \mu^{\pi}),
\end{equation*}
where $\mu^{\pi}$ is the mean field flow generated from $m_0$ when using policy $\pi$.
 Intuitively a large exploitability means that, when the population plays $\pi$, any individual player can be much better off by deviating and choosing a different strategy, so $\pi$ is far from being a Nash equilibrium policy. Conversely, an exploitability of $0$ means that $\pi$ is an MFNE policy. Similar notions are widely used in computational game theory~\citep{zinkevich2007regret,lanctot2009monte}. 

In the sequel, we consider problems for which a BR can be computed exactly given a mean-field flow. Otherwise an approximate exploitability could be used as a good proxy to assess convergence, see \textit{e.g.}~\citet{perrin2021mfgflockrl}.

\subsection{Baselines}

To assess the quality of the proposed algorithms, we consider three baselines from the literature: Banach-Picard (BP) fixed point iterations, policy iterations (PI), and Boltzmann iterations (BI).   
FP can be viewed as a modification of the first one, while OMD as a modification of the second one. They have been discussed at the beginning of Sec.~\ref{sec:fictitious-play} and Sec.~\ref{sec:background-OMD} respectively, in the exact case. Adapting them to the model-free setting with deep networks can be done in a similar way as discussed above for D-AFP and D-MOMD. See Appx.~\ref{sec:deep-rl-algos} for more details. The third baseline has been introduced recently by \citet{pmlr-cui21-approximately}.  
It consists in updating in turn the population distribution and the policy, but here the policy is computed as a weighted softmax of the optimal $Q$-values (and hence requires the resolution of an MDP at each iteration). More precisely, given a reference policy $\pi_B$, a parameter $\eta>0$, and the $Q$-function $Q^{k}$ computed at iteration $k$, the new policy is defined as:
\begin{equation*}
    \pi^k_n(a|x) = \frac{\pi_{B,n}(a|x)\exp\big(Q^k_n(x,a)/ \eta \big)}{\sum_{a'} \pi_{B,n}(a'|x)\exp\big(Q^k_n(x,a') / \eta\big)}.
\end{equation*}
In the plots, D-BP, D-AFP, D-PI, D-BI and D-MOMD refer respectively to Deep Banach-Picard iterations, Deep Average-network Fictitious Play, Deep Policy Iteration, Deep Boltzmann Iteration, and Deep Munchausen OMD.

\subsection{Numerical results}

\begin{figure}[tbh]
    \centering
    \begin{minipage}{.47\linewidth}
    \includegraphics[width=1.15\linewidth]{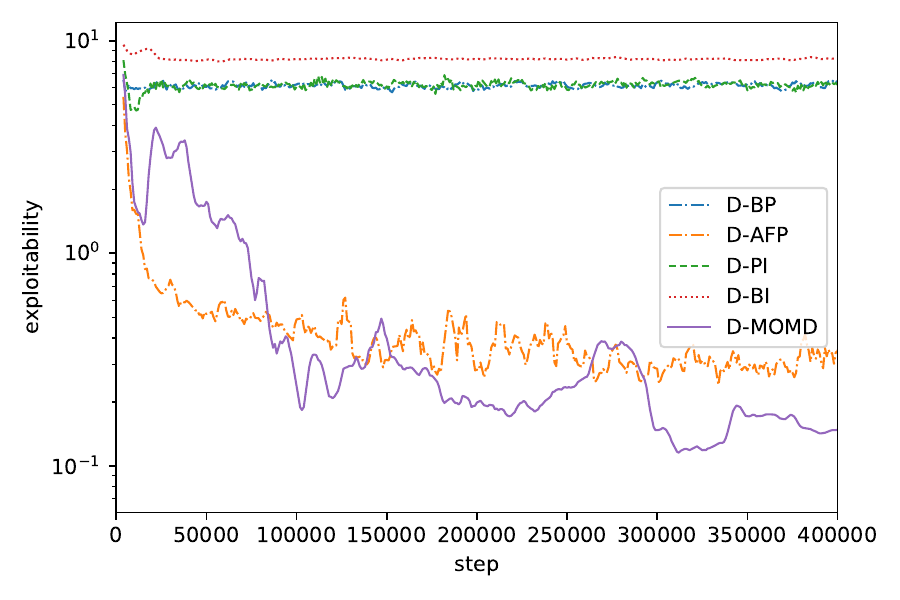} 
    \end{minipage}
    \,
    \begin{minipage}{.47\linewidth}
    \includegraphics[width=1.15\linewidth]{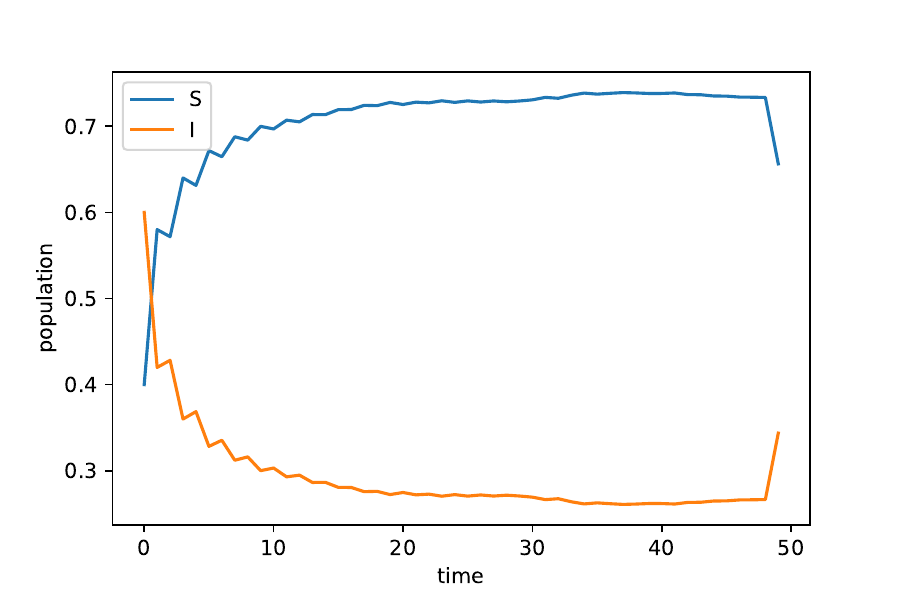} 
    \end{minipage} 

    \caption{
    Left: exploitability. Right: evolution of the distribution obtained by the policy learnt with D-MOMD.
    }
    \label{fig:sis_example}
\end{figure}

\textbf{Epidemics model. } We first consider the SIS model of~\citet{pmlr-cui21-approximately}, which is a toy model for epidemics. There are two states: susceptible (S) and infected (I). Two actions can be used: social distancing (D) or going out (U). The probability of getting infected increases if more people are infected, and is smaller when using D instead of U. The transitions are: $p(S|I,D,\mu)=p(S|I,U,\mu)=0.3$, $p(I|S,U,\mu)=0.9^2 \cdot\mu(I)$, $p(I|S,D,\mu)=0$, the reward is: $r(s,a,\mu) = -1_{I}(s) - 0.5 \cdot 1_{D}(s)$, and the horizon is $N_T=50$. Note that, although this model has only two states, the state dynamics is impacted by the distribution, which is generally challenging when computing MFG solutions.  
As shown in Fig.~\ref{fig:sis_example}, both D-MOMD and D-AFP generate an exploitability diminishing with the learning steps, whereas the other baselines are not able to do so. Besides,  D-MOMD generates smooth state trajectories, as opposed to the one observed in   \citet{pmlr-cui21-approximately}, that contained many oscillations.
For this example and the following ones, we display the best exploitability curves obtained for each method after running sweeps over hyperparameters. See Appx.~\ref{sec:hyperparams-sweeps} for some instances of sweeps for D-MOMD.

\textbf{Linear-Quadratic MFG.}
We then consider an example with more states, in 1D: the classical linear-quadratic environment of \citet{carmonafouquesun2015mean}, 
which admits an explicit closed form solution in a continuous space-time domain. We focus on a  discretized approximation  \citep{perrin2020fictitious} of the time grid $\{0,\ldots,N_T\}$, where the dynamics of the underlying state process controlled by action $a_n$ is given by
$ 
x_{n+1} = x_n + a_n\Delta_n +\sigma \epsilon_n \sqrt{\Delta_n}\;,$
with $(\bar m_n)_n$ the average of the population states, $\Delta_n$ the time step and $(\epsilon_n)_n$ i.i.d. noises on $\{-3 ,-2,-1,0,1,2,3\}$, truncated approximations of $\mathcal{N}(0,1)$ random variables. Given a set of actions $(a_n)_n$ in state trajectory $(x_n)_n$ and a mean field flow $(\mu_n)_n$, the reward $r(x_n,a_n,\mu_n)$ of a representative player is given for $n< N_T$ by
 $$
  \Big[ -\frac{1}{2} |a_n|^2 + q a_n (\bar m_n-x_n) -\frac{\kappa}{2} |\bar m_n-x_n|^2 \Big]\Delta_n\,,
 $$
 together with the terminal reward $-\frac{c_{term}}{2} |\bar m_{N_T}-x_{N_T}|^2$. The reward penalizes high actions, while providing incentives to remain close to the average state despite the noise $(\epsilon_n)_n$. 
 For the experiments, we used $N_T=10$,  $\sigma=1$, $\Delta_n=1$, $q=0.01$, $\kappa=0.5$, $c_{term}=1$ and $\lvert \states \rvert= 100$. The action space is $\{-3, -2, -1, 0, 1, 2, 3\}$. 
 
\begin{figure}[tbh]
    \centering
    \begin{minipage}{.62\linewidth}
    \includegraphics[width=1.1\linewidth]{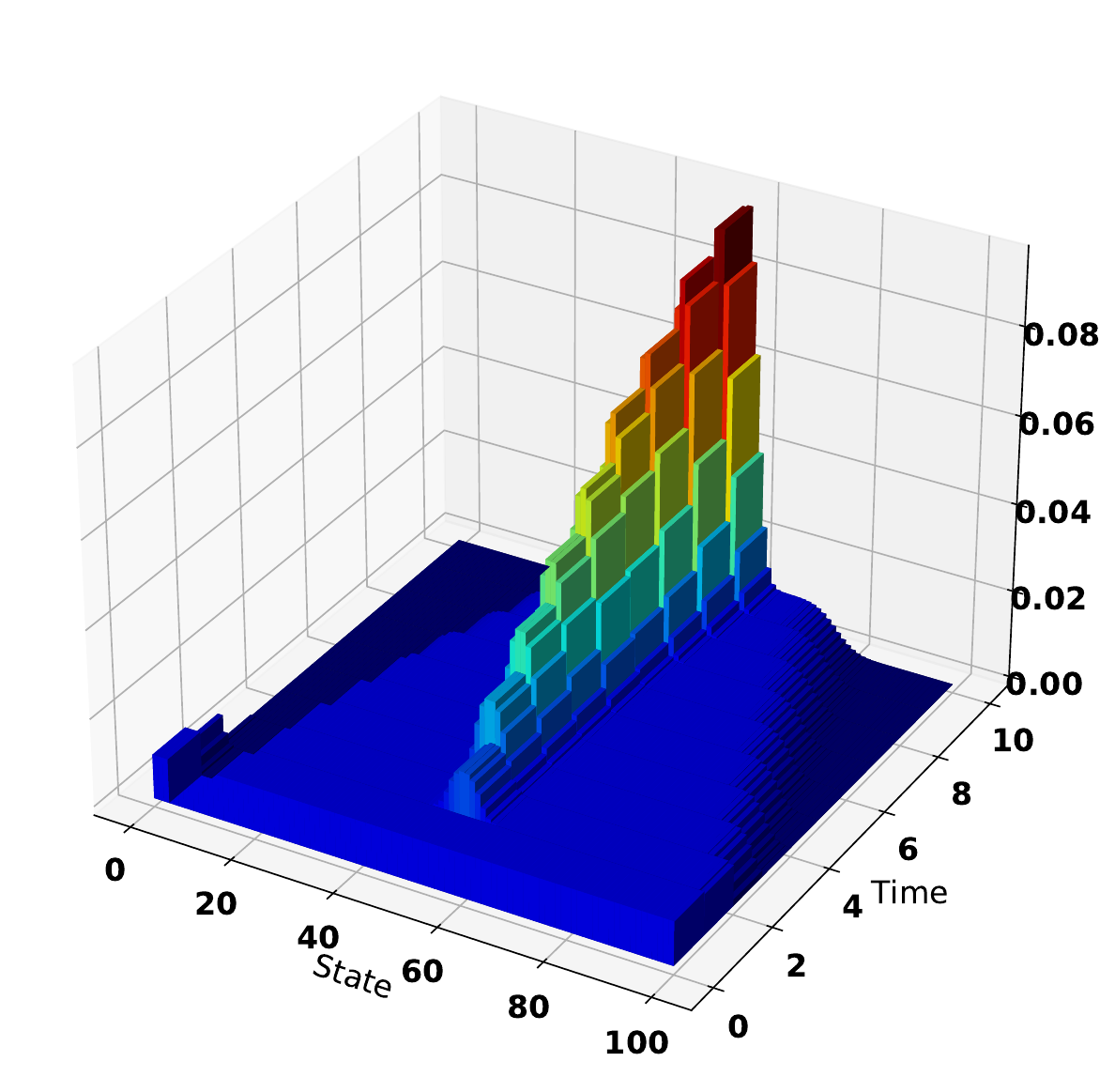}    
    \end{minipage}
    \hfill
    \begin{minipage}{.42\linewidth}\hspace{-0.5cm}
    \includegraphics[width=1.1\linewidth]{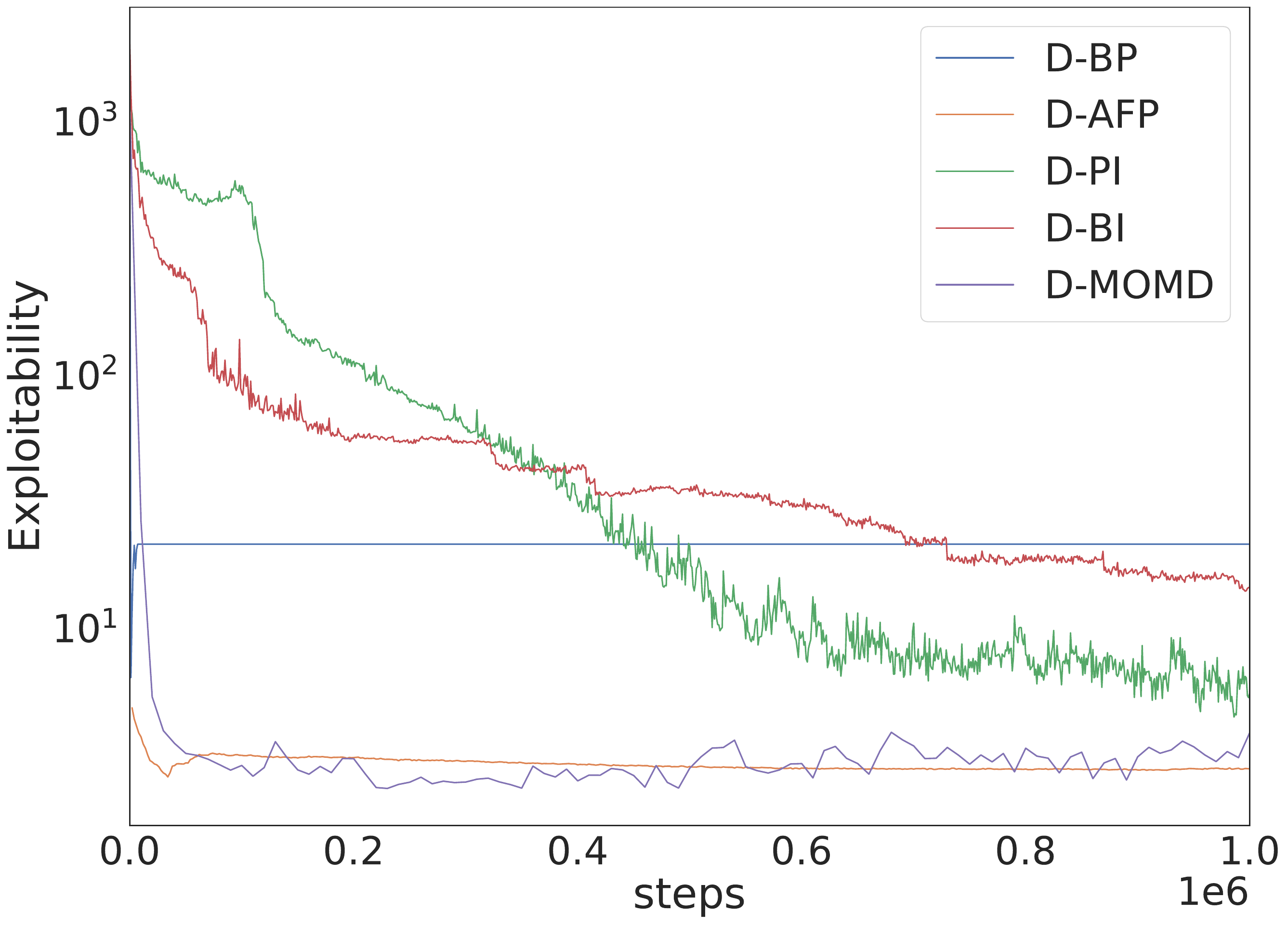}    
    \end{minipage}
    \begin{minipage}{.42\linewidth}
    \includegraphics[width=1.1\linewidth]{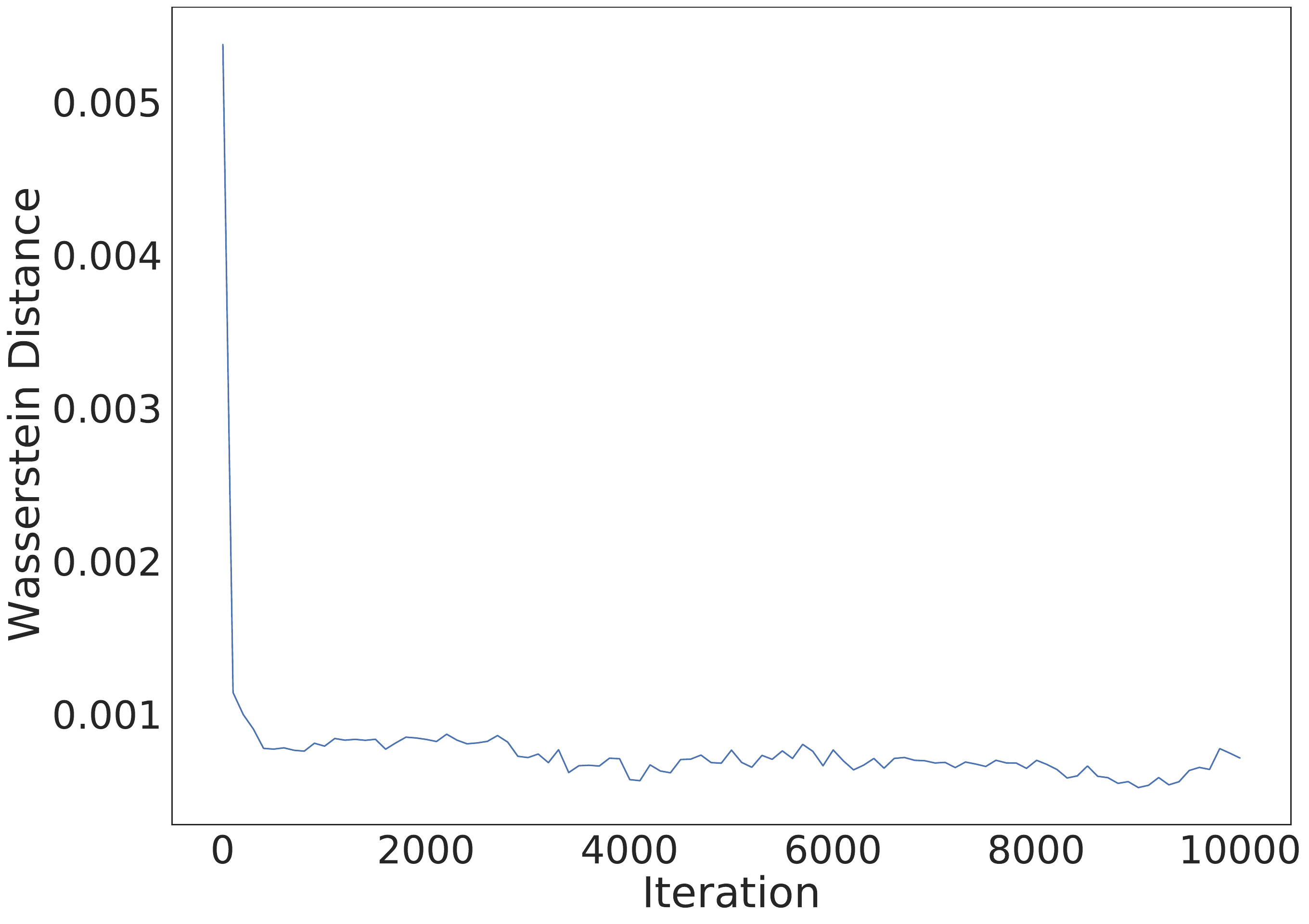}
    \end{minipage}

    \caption{ 
    Top: Evolution of the distribution generated by the policy learnt by D-MOMD.
    Bottom left: Exploitability of different algorithms on the Linear Quadratic environment. Bottom right: Wasserstein distance between the solution learnt by D-MOMD and the benchmark one, over its iterations.}
    \label{fig:lq_example}
\end{figure}

  In Figure \ref{fig:lq_example} (top), we see that the distribution estimated by D-MOMD concentrates, as is expected from the reward encouraging a mean-reverting behavior: the population gathers as expected into a bell-shaped distribution. The  analytical solution (Appx. E in \citep{perrin2020fictitious}) is exact in a continuous setting (\textit{i.e.}, when the step sizes in time, state and action go to $0$) but only approximate in the discrete one considered here. Hence, we choose instead to use the distribution estimated by exact tabular OMD as a benchmark, as it reaches an exploitability of $10^{-12}$ in our experiments. Figure \ref{fig:lq_example} (bottom right) shows that the Wasserstein distance between the learnt distribution by D-MOMD and the benchmark decreases as the learning occurs. In Figure \ref{fig:lq_example} (bottom left), we see that D-MOMD and D-AFP outperform other methods in minimizing exploitability. 

\begin{figure}[tbh]
    \centering
    \begin{minipage}{.7\linewidth}
    \includegraphics[width=.85\linewidth]{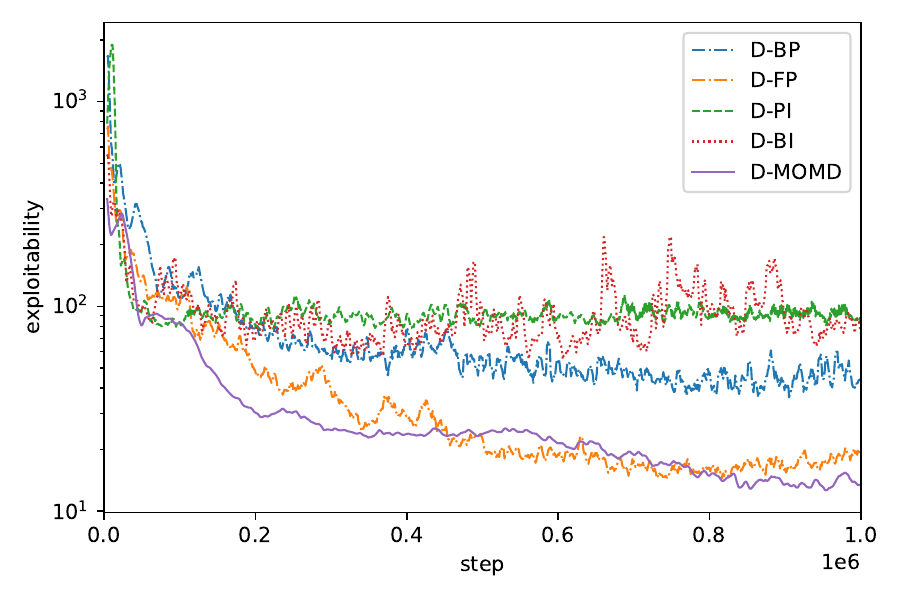} 
    \end{minipage}
    
    \begin{minipage}{\linewidth}
    \includegraphics[width=\linewidth]{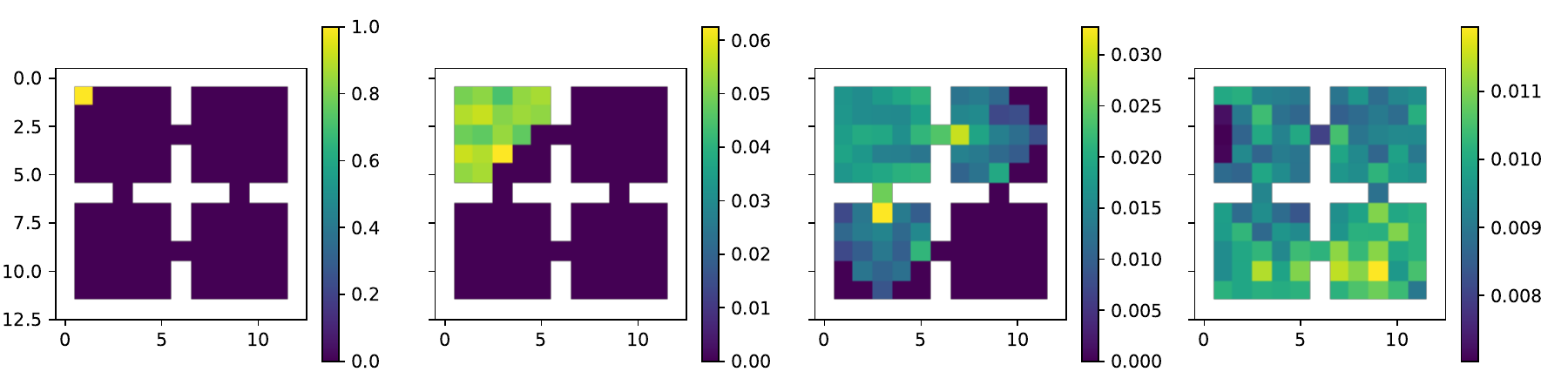}
    \end{minipage}
    \caption{ 
    Top: exploitability. Bottom: evolution of the distribution obtained by the policy learnt with D-MOMD.    
    }
    \label{fig:curl_example}
\end{figure}

\textbf{Exploration. }
\looseness=-1
We now increase the state dimension and turn our attention to a 2-dimensional grid world example. The state is the position. An action is a move, and valid moves are: left, right, up, down, or stay, as long as the agent does not hit a wall. In the experiments, we consider $10 \times 10$ states and a time horizon of $N_T=40$ time steps. The reward is: 
$
    r(x,a,\mu) = r_{\mathrm{pop}}(\mu(x)),
$ 
where $r_{\mathrm{pop}}(\mu(x)) = -\log(\mu(x))$ discourages being in a crowded area -- which is referred to as crowd aversion. Note that $\EE_{x \sim \mu}(r_{\mathrm{pop}}(\mu(x))) = \mathcal{H}(\mu)$, \textit{i.e.}, the last term of the reward provides, in expectation, the entropy of the distribution. This setting is inspired by the one considered by \citet{geist2021curl}.  The results are shown in Fig.~\ref{fig:curl_example}. D-MOMD and D-AFP outperform all the baselines.  The induced distribution matches our intuition: it spreads symmetrically until filling almost uniformly the four rooms. 

\begin{figure}[tbh]
    \centering
    \begin{minipage}{.7\linewidth}
    \includegraphics[width=.85\linewidth]{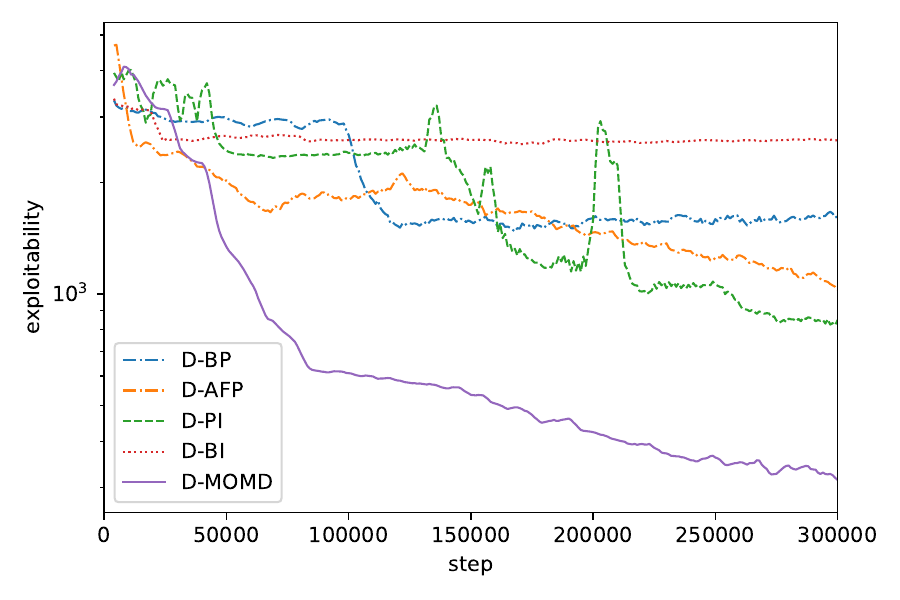} 
    \end{minipage}
    
    \begin{minipage}{\linewidth}
    \includegraphics[width=\linewidth]{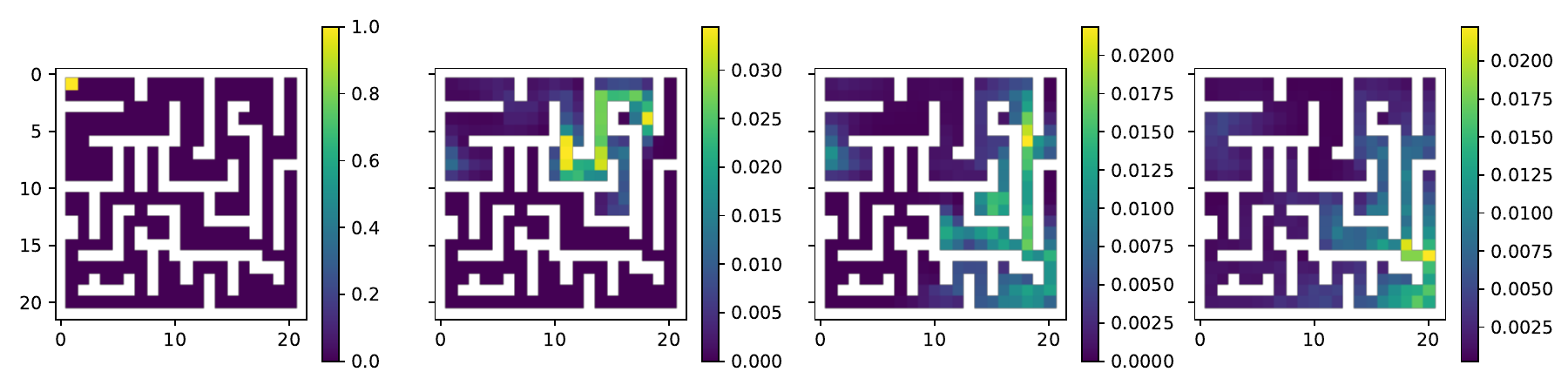}
    \end{minipage}
    \caption{ 
    Maze example. Top: exploitability. Bottom: evolution of the distribution obtained by the policy learnt with D-MOMD.  
    }
    \label{fig:maze_example}
\end{figure}

\textbf{Crowd modeling with congestion. }
We consider the same environment but with a maze, and a more complex reward function:
\begin{equation*}
    r(x,a,\mu) = r_{\mathrm{pos}}(x) + r_{\mathrm{move}}(a,\mu(x)) + r_{\mathrm{pop}}(\mu(x)),
\end{equation*}
\looseness=-1
where $r_{\mathrm{pos}}(x) = - \mathrm{dist}(x, x_{\mathrm{ref}})$ is the distance to a target position $x_{\mathrm{ref}}$, $r_{\mathrm{move}}(a,\mu(x)) = -\mu(x)\|a\|$ is a penalty for moving ($\|a\|=1$) which increases with the density $\mu(x)$ at $x$ -- which is called congestion effect in the literature. The state space has $20 \times 20 $ states, and the time horizon is $N_T=100$. We see in Fig.~\ref{fig:maze_example} that D-MOMD outperforms the other methods.

\begin{figure}[tbh]
    \centering
    \begin{minipage}{.7\linewidth}
    \includegraphics[width=\linewidth]{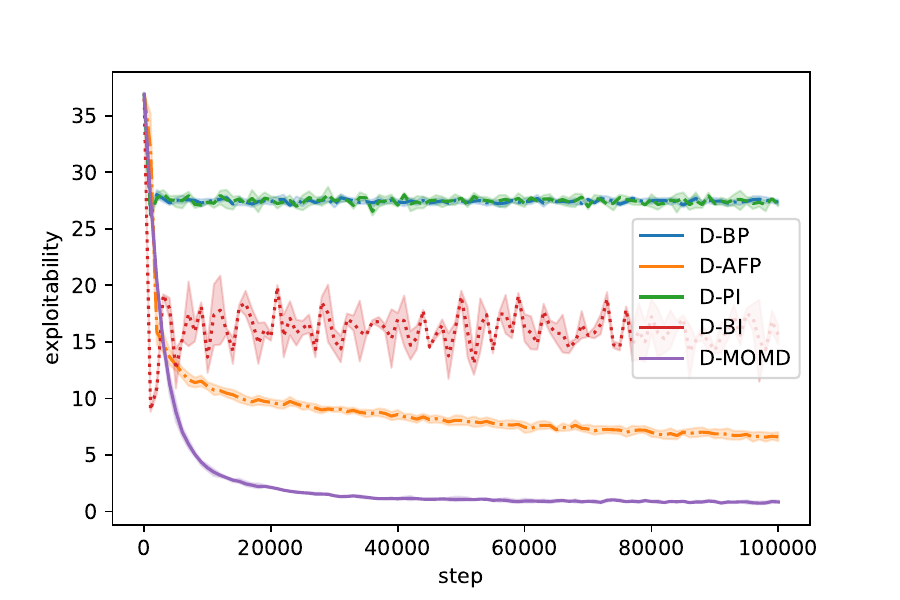} 
    \end{minipage}
    
    \begin{minipage}{.7\linewidth}
    \includegraphics[width=\linewidth]{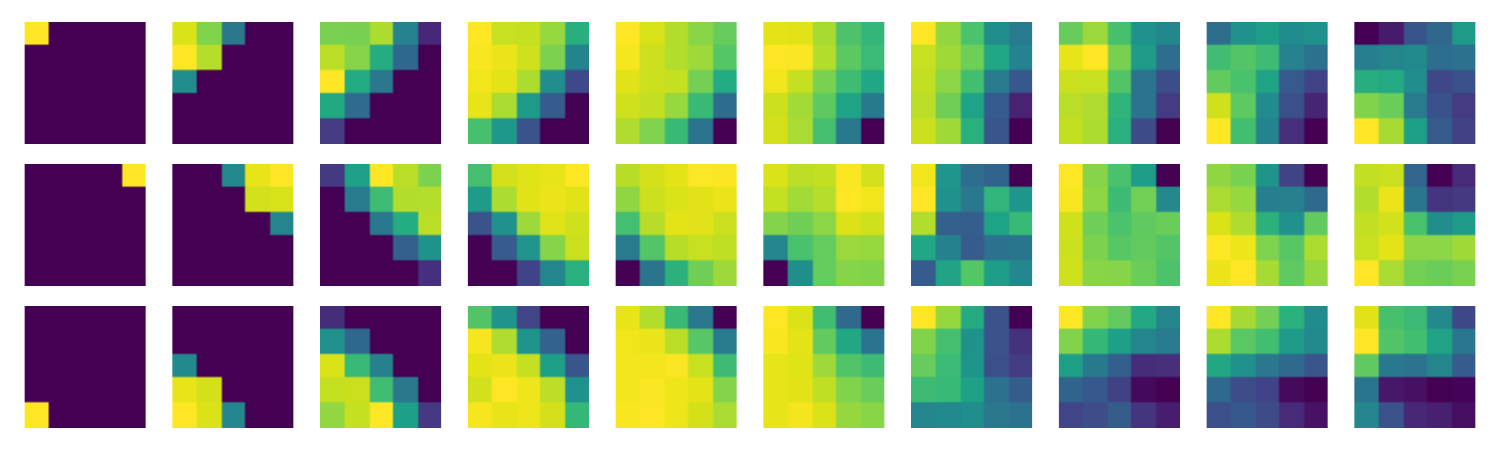} 
    \end{minipage}
    
    \caption{
    Multi-population chasing example. Top: Exploitability. Bottom: evolution of the distributions for the three populations.
    }
    \label{fig:predator_prey_example}
\end{figure}

\textbf{Multi-population chasing. } 
We finally turn to an extension of the MFG framework, where agents are heterogeneous: each type of agent has its own dynamics and reward function. The environment can be extended to model multiple populations by simply extending the state space to include the population index on top of the agent's position. Following \citet{perolat2021scaling}, we consider three populations and rewards of the form: for population $i=1,2,3$,
$$
    r^i(x, a, \mu^1, \mu^2, \mu^3) = -\log(\mu^i(x)) + \sum_{j\neq i}\mu^j(x)\bar r^{i,j}(x).
$$
where $\bar r^{i,j} = - \bar r^{j,i}$, with $\bar r^{1,2} = -1,$ $\bar r^{1,3}=1,$ $\bar r^{2,3} = -1$.
In the experiments, three are $5 \times 5$ states and the time horizon is $N_T=10$ time steps. The initial distributions are in the corners, the number of agents of each population is fixed, and the reward encourages the agent to chase the population it dominates and flee the dominating one. We see in Fig.~\ref{fig:predator_prey_example} that D-AFP outperform the baselines and D-MOMD performs even better. 

\section{Conclusion}

In this work, we proposed two scalable algorithms that can compute Nash equilibria in various MFGs in the finite horizon setting. The first one, D-AFP, is the first implementation of Fictitious Play for MFGs that does not need to keep all previous best responses in memory and that learns an average policy with a single neural network. The second one, D-MOMD, takes advantage of a subtle reparameterization to learn implicitly a sum of $Q$-functions usually required in the Online Mirror Descent algorithm. We demonstrated numerically that they both perform well on five benchmark problems and that D-MOMD consistently performs better than D-AFP as well as three baselines from the literature.

\paragraph{Related Work.} Only a few works try to learn Nash equilibria in non-stationary MFGs. \citet{guohuxuzhang2020generalmfg} establish the existence and uniqueness of dynamic Nash equilibrium under strict regularity assumptions in their Section 9, but their RL and numerical experiments are restricted to the stationary setting. \citet{perolat2021scaling} prove that continuous-time OMD converges to the Nash equilibrium in monotone MFGs, supported by numerical experiments involving trillions of states; however, their approach is purely based on the model and not RL. \citet{mishra2020modelfreenonstatmfg} propose an RL-based sequential approach to dynamic MFG, but this method relies on functions of the distribution, which is not scalable when the state space is large.  \citet{perrin2020fictitious} prove the convergence of FP for monotone MFGs and provide experimental results with model-free methods ($Q$-learning); however, they avoid the difficulty of mixing policies by relying on tabular representations for small environments. \citet{pmlr-cui21-approximately} propose a deep RL method with regularization to learn Nash equilibria in finite horizon MFGs; however, the regularization biases the equilibrium and they were not able to make it converge in complex examples, whereas D-MOMD outperforms the baselines in all the games we considered. \citet{perrin2021generalizationmfg} and \citet{perrin2021mfgflockrl} use another version of FP with neural networks, but their approach needs to keep in memory all the BRs learnt during the iterations, which cannot scale to complex MFGs. 
In our D-OMD algorithm the policy is computed using a softmax, which is reminiscent of maximum entropy RL methods~\citep{todorov2008general,toussaint2009robot,rawlik2012stochastic} which led to efficient deep RL methods such as soft actor critic~\citep{haarnoja2018soft}. However there is a crucial difference: here we use a KL divergence with respect to the previous policy, as can be seen in~\eqref{eq:omd-KL-reformulation}.

\looseness=-1

\paragraph{Future work.} We would like to include more complex examples, with larger state spaces or even continuous spaces. Continuous state spaces should be relatively easy to address, as neural networks can handle continuous inputs, while continuous actions would require some adjustments particularly to compute argmax or softmax of $Q$-functions. Furthermore continuous spaces require manipulating continuous population distributions, raising additional questions related to how to represent and estimate them efficiently.

\section*{Acknowledgements}

We would like to thank the anonymous referees for their helpful comments. This project was done while Mathieu Lauri{\`e}re was a visiting faculty researcher at Google Brain. This research project is supported in part by the National Research Foundation, Singapore under its AI Singapore Program (AISG Award No: AISG2-RP-2020-016), NRF 2018 Fellowship
NRF-NRFF2018-07, NRF2019-NRF-ANR095 ALIAS grant, grant PIE-SGP-AI-2020-01, AME Programmatic Fund (Grant No. A20H6b0151) from the Agency for Science, Technology and Research
(A*STAR) and Provost’s Chair Professorship grant RGEPPV2101.

\bibliographystyle{plainnat}
\bibliography{bib}

\begin{thebibliography}{49}
\providecommand{\natexlab}[1]{#1}
\providecommand{\url}[1]{\texttt{#1}}
\expandafter\ifx\csname urlstyle\endcsname\relax
  \providecommand{\doi}[1]{doi: #1}\else
  \providecommand{\doi}{doi: \begingroup \urlstyle{rm}\Url}\fi

\bibitem[Achdou and Capuzzo-Dolcetta(2010)]{MR2679575}
Yves Achdou and Italo Capuzzo-Dolcetta.
\newblock Mean field games: numerical methods.
\newblock \emph{SIAM J. Numer. Anal.}, 48\penalty0 (3):\penalty0 1136--1162,
  2010.

\bibitem[Achdou and
  Lauri{\`e}re(2020)]{achdou2020meanfieldgamesnumericalcetraro}
Yves Achdou and Mathieu Lauri{\`e}re.
\newblock Mean field games and applications: Numerical aspects.
\newblock \emph{Mean Field Games}, pages 249--307, 2020.

\bibitem[Anahtarci et~al.(2022)Anahtarci, Kariksiz, and
  Saldi]{anahtarci2020qregmfg}
Berkay Anahtarci, Can~Deha Kariksiz, and Naci Saldi.
\newblock Q-learning in regularized mean-field games.
\newblock \emph{Dynamic Games and Applications}, pages 1--29, 2022.

\bibitem[Brice\~{n}o Arias et~al.(2018)Brice\~{n}o Arias, Kalise, and
  Silva]{MR3772008}
Luis~M. Brice\~{n}o Arias, Dante Kalise, and Francisco~J. Silva.
\newblock Proximal methods for stationary mean field games with local
  couplings.
\newblock \emph{SIAM J. Control Optim.}, 56\penalty0 (2):\penalty0 801--836,
  2018.

\bibitem[Brown(1951)]{brown1951iterative}
George~W Brown.
\newblock Iterative solution of games by fictitious play.
\newblock \emph{Activity analysis of production and allocation}, 13\penalty0
  (1):\penalty0 374--376, 1951.

\bibitem[Brown and Sandholm(2018)]{brown2018superhuman}
Noam Brown and Tuomas Sandholm.
\newblock Superhuman ai for heads-up no-limit poker: Libratus beats top
  professionals.
\newblock \emph{Science}, 359\penalty0 (6374):\penalty0 418--424, 2018.

\bibitem[Cacace et~al.(2021)Cacace, Camilli, and Goffi]{cacace2020policy}
Simone Cacace, Fabio Camilli, and Alessandro Goffi.
\newblock A policy iteration method for mean field games.
\newblock \emph{ESAIM: COCV}, 27:\penalty0 85, 2021.

\bibitem[Campbell et~al.(2002)Campbell, Hoane~Jr, and Hsu]{campbell2002deep}
Murray Campbell, A~Joseph Hoane~Jr, and Feng-hsiung Hsu.
\newblock Deep blue.
\newblock \emph{Artificial intelligence}, 134\penalty0 (1-2):\penalty0 57--83,
  2002.

\bibitem[Cao et~al.(2020)Cao, Guo, and Lauri{\`e}re]{cao2020connecting}
Haoyang Cao, Xin Guo, and Mathieu Lauri{\`e}re.
\newblock Connecting {GAN}s, mean-field games, and optimal transport.
\newblock \emph{arXiv preprint arXiv:2002.04112}, 2020.

\bibitem[Cardaliaguet and Hadikhanloo(2017)]{hadikhanloo_fictitious-play}
Pierre Cardaliaguet and Saeed Hadikhanloo.
\newblock Learning in mean field games: the fictitious play.
\newblock \emph{ESAIM Cont. Optim. Calc. Var.}, 2017.

\bibitem[Carlini and Silva(2014)]{MR3148086}
Elisabetta Carlini and Francisco~J. Silva.
\newblock A fully discrete semi-{L}agrangian scheme for a first order mean
  field game problem.
\newblock \emph{SIAM J. Numer. Anal.}, 52\penalty0 (1):\penalty0 45--67, 2014.

\bibitem[Carmona and Lauri{\`e}re(2019)]{carmona2019convergencefinite}
Ren{\'e} Carmona and Mathieu Lauri{\`e}re.
\newblock Convergence analysis of machine learning algorithms for the numerical
  solution of mean field control and games: {II}--the finite horizon case.
\newblock \emph{Forthcoming in Annals of Applied Probability
  (arXiv:1908.01613)}, 2019.

\bibitem[Carmona and Lauri{\`e}re(2022)]{carmonalauriere2021deepmfgsurvey}
Ren{\'e} Carmona and Mathieu Lauri{\`e}re.
\newblock Deep learning for mean field games and mean field control with
  applications to finance.
\newblock \emph{Machine Learning in Financial Markets: A guide to contemporary
  practises, editors: A. Capponi and C.-A. Lehalle, Cambridge University Press
  (Preprint arXiv:2107.04568)}, 2022.

\bibitem[Carmona et~al.(2015)Carmona, Fouque, and
  Sun]{carmonafouquesun2015mean}
Ren{\'e} Carmona, Jean-Pierre Fouque, and Li-Hsien Sun.
\newblock Mean field games and systemic risk.
\newblock \emph{Communications in Mathematical Sciences}, 13\penalty0
  (4):\penalty0 911--933, 2015.

\bibitem[Cui and Koeppl(2021)]{pmlr-cui21-approximately}
Kai Cui and Heinz Koeppl.
\newblock Approximately solving mean field games via entropy-regularized deep
  reinforcement learning.
\newblock In \emph{proc. of AISTATS}, 2021.

\bibitem[Daskalakis et~al.(2006)Daskalakis, Goldberg, and
  Papadimitriou]{daskalakis2008}
Constantinos Daskalakis, Paul~W. Goldberg, and Christos~H. Papadimitriou.
\newblock The complexity of computing a {N}ash equilibrium.
\newblock In \emph{proc. of ACM Symposium on Theory of Computing}, 2006.

\bibitem[Elie et~al.(2020)Elie, Perolat, Lauri{\`e}re, Geist, and
  Pietquin]{elie2020convergence}
Romuald Elie, Julien Perolat, Mathieu Lauri{\`e}re, Matthieu Geist, and Olivier
  Pietquin.
\newblock On the convergence of model free learning in mean field games.
\newblock In \emph{proc. of AAAI}, 2020.

\bibitem[Fouque and Zhang(2020)]{fouque2020deep}
Jean-Pierre Fouque and Zhaoyu Zhang.
\newblock Deep learning methods for mean field control problems with delay.
\newblock \emph{Frontiers in Applied Mathematics and Statistics}, 6:\penalty0
  11, 2020.

\bibitem[Geist et~al.(2022)Geist, P{\'e}rolat, Lauri{\`e}re, Elie, Perrin,
  Bachem, Munos, and Pietquin]{geist2021curl}
Matthieu Geist, Julien P{\'e}rolat, Mathieu Lauri{\`e}re, Romuald Elie, Sarah
  Perrin, Olivier Bachem, R{\'e}mi Munos, and Olivier Pietquin.
\newblock Concave utility reinforcement learning: the mean-field game
  viewpoint.
\newblock In \emph{proc. of AAMAS}, 2022.

\bibitem[Germain et~al.(2022)Germain, Mikael, and Warin]{germain2019numerical}
Maximilien Germain, Joseph Mikael, and Xavier Warin.
\newblock Numerical resolution of {M}c{K}ean-{V}lasov {FBSDE}s using neural
  networks.
\newblock \emph{Methodology and Computing in Applied Probability}, pages 1--30,
  2022.

\bibitem[Goodfellow et~al.(2016)Goodfellow, Bengio, and
  Courville]{goodfellow2016deep}
Ian Goodfellow, Yoshua Bengio, and Aaron Courville.
\newblock \emph{Deep learning}.
\newblock MIT press, 2016.

\bibitem[Goodfellow et~al.(2020)Goodfellow, Pouget-Abadie, Mirza, Xu,
  Warde-Farley, Ozair, Courville, and Bengio]{goodfellow2020generative}
Ian Goodfellow, Jean Pouget-Abadie, Mehdi Mirza, Bing Xu, David Warde-Farley,
  Sherjil Ozair, Aaron Courville, and Yoshua Bengio.
\newblock Generative adversarial networks.
\newblock \emph{Communications of the ACM}, 2020.

\bibitem[Guo et~al.(2020)Guo, Hu, Xu, and Zhang]{guohuxuzhang2020generalmfg}
Xin Guo, Anran Hu, Renyuan Xu, and Junzi Zhang.
\newblock A general framework for learning mean-field games.
\newblock \emph{arXiv preprint arXiv:2003.06069}, 2020.

\bibitem[Guo et~al.(2022)Guo, Xu, and
  Zariphopoulou]{guoxuzariphopoulou2020entropyregmfg}
Xin Guo, Renyuan Xu, and Thaleia Zariphopoulou.
\newblock Entropy regularization for mean field games with learning.
\newblock \emph{Mathematics of Operations Research}, 2022.

\bibitem[Haarnoja et~al.(2018)Haarnoja, Zhou, Abbeel, and
  Levine]{haarnoja2018soft}
Tuomas Haarnoja, Aurick Zhou, Pieter Abbeel, and Sergey Levine.
\newblock Soft actor-critic: Off-policy maximum entropy deep reinforcement
  learning with a stochastic actor.
\newblock In \emph{International conference on machine learning}, pages
  1861--1870. PMLR, 2018.

\bibitem[Hadikhanloo(2017)]{hadikhanloo2017learningnonatomic}
Saeed Hadikhanloo.
\newblock Learning in anonymous nonatomic games with applications to
  first-order mean field games.
\newblock \emph{arXiv preprint arXiv:1704.00378}, 2017.

\bibitem[Heinrich and Silver(2016)]{heinrichsilver2016deepnfsp}
Johannes Heinrich and David Silver.
\newblock Deep reinforcement learning from self-play in imperfect-information
  games.
\newblock \emph{arXiv preprint arXiv:1603.01121}, 2016.

\bibitem[Huang et~al.(2006)Huang, Malhamé, and Caines]{2006huang-mfg}
Minyi Huang, Roland~P. Malhamé, and Peter~E. Caines.
\newblock {Large population stochastic dynamic games: closed-loop McKean-Vlasov
  systems and the {N}ash certainty equivalence principle}.
\newblock \emph{Communications in Information \& Systems}, 6, 2006.

\bibitem[Lanctot et~al.(2009)Lanctot, Waugh, Zinkevich, and
  Bowling]{lanctot2009monte}
Marc Lanctot, Kevin Waugh, Martin Zinkevich, and Michael Bowling.
\newblock Monte {C}arlo sampling for regret minimization in extensive games.
\newblock In \emph{proc. of NeurIPS}, volume~22, pages 1078--1086, 2009.

\bibitem[Lanctot et~al.(2017)Lanctot, Zambaldi, Gruslys, Lazaridou, Tuyls,
  P{\'e}rolat, Silver, and Graepel]{lanctot2017unified}
Marc Lanctot, Vinicius Zambaldi, Audrunas Gruslys, Angeliki Lazaridou, Karl
  Tuyls, Julien P{\'e}rolat, David Silver, and Thore Graepel.
\newblock A unified game-theoretic approach to multiagent reinforcement
  learning.
\newblock \emph{Advances in neural information processing systems}, 30, 2017.

\bibitem[Lanctot et~al.(2019)Lanctot, Lockhart, Lespiau, Zambaldi, Upadhyay,
  P{\'e}rolat, Srinivasan, Timbers, Tuyls, Omidshafiei,
  et~al.]{lanctot2019openspiel}
Marc Lanctot, Edward Lockhart, Jean-Baptiste Lespiau, Vinicius Zambaldi,
  Satyaki Upadhyay, Julien P{\'e}rolat, Sriram Srinivasan, Finbarr Timbers,
  Karl Tuyls, Shayegan Omidshafiei, et~al.
\newblock Openspiel: A framework for reinforcement learning in games.
\newblock \emph{arXiv preprint arXiv:1908.09453}, 2019.

\bibitem[Lasry and Lions(2007)]{MFG_lasry-lions}
Jean-Michel Lasry and Pierre-Louis Lions.
\newblock Mean field games.
\newblock \emph{Jpn. J. Math.}, 2007.
\newblock ISSN 0289-2316.

\bibitem[Mishra et~al.(2020)Mishra, Vasal, and
  Vishwanath]{mishra2020modelfreenonstatmfg}
Rajesh~K Mishra, Deepanshu Vasal, and Sriram Vishwanath.
\newblock Model-free reinforcement learning for non-stationary mean field
  games.
\newblock In \emph{2020 59th IEEE Conference on Decision and Control (CDC)},
  pages 1032--1037. IEEE, 2020.

\bibitem[Mnih et~al.(2013)Mnih, Kavukcuoglu, Silver, Graves, Antonoglou,
  Wierstra, and Riedmiller]{mnih2013playingatari}
Volodymyr Mnih, Koray Kavukcuoglu, David Silver, Alex Graves, Ioannis
  Antonoglou, Daan Wierstra, and Martin Riedmiller.
\newblock Playing atari with deep reinforcement learning.
\newblock \emph{arXiv preprint arXiv:1312.5602}, 2013.

\bibitem[Morav{\v{c}}{\'\i}k et~al.(2017)Morav{\v{c}}{\'\i}k, Schmid, Burch,
  Lis{\`y}, Morrill, Bard, Davis, Waugh, Johanson, and
  Bowling]{moravvcik2017deepstack}
Matej Morav{\v{c}}{\'\i}k, Martin Schmid, Neil Burch, Viliam Lis{\`y}, Dustin
  Morrill, Nolan Bard, Trevor Davis, Kevin Waugh, Michael Johanson, and Michael
  Bowling.
\newblock Deepstack: Expert-level artificial intelligence in heads-up no-limit
  poker.
\newblock \emph{Science}, 356\penalty0 (6337):\penalty0 508--513, 2017.

\bibitem[Perolat et~al.(2021)Perolat, Munos, Lespiau, Omidshafiei, Rowland,
  Ortega, Burch, Anthony, Balduzzi, De~Vylder, et~al.]{perolat2021poincare}
Julien Perolat, Remi Munos, Jean-Baptiste Lespiau, Shayegan Omidshafiei, Mark
  Rowland, Pedro Ortega, Neil Burch, Thomas Anthony, David Balduzzi, Bart
  De~Vylder, et~al.
\newblock From poincar{\'e} recurrence to convergence in imperfect information
  games: Finding equilibrium via regularization.
\newblock In \emph{proc. of ICML}, 2021.

\bibitem[Perolat et~al.(2022)Perolat, Perrin, Elie, Lauri{\`e}re, Piliouras,
  Geist, Tuyls, and Pietquin]{perolat2021scaling}
Julien Perolat, Sarah Perrin, Romuald Elie, Mathieu Lauri{\`e}re, Georgios
  Piliouras, Matthieu Geist, Karl Tuyls, and Olivier Pietquin.
\newblock Scaling up mean field games with online mirror descent.
\newblock In \emph{proc. of AAAI}, 2022.

\bibitem[Perrin et~al.(2020)Perrin, P{\'e}rolat, Lauri{\`e}re, Geist, Elie, and
  Pietquin]{perrin2020fictitious}
Sarah Perrin, Julien P{\'e}rolat, Mathieu Lauri{\`e}re, Matthieu Geist, Romuald
  Elie, and Olivier Pietquin.
\newblock Fictitious play for mean field games: Continuous time analysis and
  applications.
\newblock In \emph{proc. of NeurIPS}, 2020.

\bibitem[Perrin et~al.(2021)Perrin, Lauri{\`e}re, P{\'e}rolat, Geist, {\'E}lie,
  and Pietquin]{perrin2021mfgflockrl}
Sarah Perrin, Mathieu Lauri{\`e}re, Julien P{\'e}rolat, Matthieu Geist, Romuald
  {\'E}lie, and Olivier Pietquin.
\newblock Mean field games flock! {T}he reinforcement learning way.
\newblock In \emph{proc. of IJCAI}, 2021.

\bibitem[Perrin et~al.(2022)Perrin, Lauri{\`e}re, P{\'e}rolat, {\'E}lie, Geist,
  and Pietquin]{perrin2021generalizationmfg}
Sarah Perrin, Mathieu Lauri{\`e}re, Julien P{\'e}rolat, Romuald {\'E}lie,
  Matthieu Geist, and Olivier Pietquin.
\newblock Generalization in mean field games by learning master policies.
\newblock In \emph{proc. of AAAI}, 2022.

\bibitem[Rawlik et~al.(2012)Rawlik, Toussaint, and
  Vijayakumar]{rawlik2012stochastic}
Konrad Rawlik, Marc Toussaint, and Sethu Vijayakumar.
\newblock On stochastic optimal control and reinforcement learning by
  approximate inference.
\newblock \emph{Proceedings of Robotics: Science and Systems VIII}, 2012.

\bibitem[Silver et~al.(2016)Silver, Huang, Maddison, Guez, Sifre, Van
  Den~Driessche, Schrittwieser, Antonoglou, Panneershelvam, Lanctot,
  et~al.]{silver2016mastering}
David Silver, Aja Huang, Chris~J Maddison, Arthur Guez, Laurent Sifre, George
  Van Den~Driessche, Julian Schrittwieser, Ioannis Antonoglou, Veda
  Panneershelvam, Marc Lanctot, et~al.
\newblock Mastering the game of {Go} with deep neural networks and tree search.
\newblock \emph{Nature}, 529\penalty0 (7587), 2016.

\bibitem[Todorov(2008)]{todorov2008general}
Emanuel Todorov.
\newblock General duality between optimal control and estimation.
\newblock In \emph{2008 47th IEEE Conference on Decision and Control}, pages
  4286--4292. IEEE, 2008.

\bibitem[Toussaint(2009)]{toussaint2009robot}
Marc Toussaint.
\newblock Robot trajectory optimization using approximate inference.
\newblock In \emph{Proceedings of the 26th annual international conference on
  machine learning}, pages 1049--1056, 2009.

\bibitem[Tuyls and Weiss(2012)]{Tuyls_Weiss_2012}
Karl Tuyls and Gerhard Weiss.
\newblock Multiagent learning: Basics, challenges, and prospects.
\newblock \emph{AI Magazine}, 2012.

\bibitem[Vieillard et~al.(2020)Vieillard, Pietquin, and
  Geist]{vieillard_munchausen_neurips_2020}
Nino Vieillard, Olivier Pietquin, and Matthieu Geist.
\newblock Munchausen reinforcement learning.
\newblock In \emph{proc. of NeurIPS}, 2020.

\bibitem[Vinyals et~al.(2019)Vinyals, Babuschkin, Czarnecki, Mathieu, Dudzik,
  Chung, Choi, Powell, Ewalds, Georgiev, et~al.]{vinyals2019grandmaster}
Oriol Vinyals, Igor Babuschkin, Wojciech~M Czarnecki, Micha{\"e}l Mathieu,
  Andrew Dudzik, Junyoung Chung, David~H Choi, Richard Powell, Timo Ewalds,
  Petko Georgiev, et~al.
\newblock Grandmaster level in starcraft ii using multi-agent reinforcement
  learning.
\newblock \emph{Nature}, 575\penalty0 (7782):\penalty0 350--354, 2019.

\bibitem[Xie et~al.(2021)Xie, Yang, Wang, and Minca]{pmlr-xie21-learning}
Qiaomin Xie, Zhuoran Yang, Zhaoran Wang, and Andreea Minca.
\newblock Learning while playing in mean-field games: Convergence and
  optimality.
\newblock In Marina Meila and Tong Zhang, editors, \emph{Proceedings of the
  38th International Conference on Machine Learning}, volume 139, pages
  11436--11447. PMLR, 2021.

\bibitem[Zinkevich et~al.(2007)Zinkevich, Johanson, Bowling, and
  Piccione]{zinkevich2007regret}
Martin Zinkevich, Michael Johanson, Michael Bowling, and Carmelo Piccione.
\newblock Regret minimization in games with incomplete information.
\newblock In \emph{proc. of NeurIPS}, 2007.

\end{thebibliography}

\newpage
\appendix
\onecolumn

\section{Algorithms in the exact case}
\label{sec:algo-exact}

In this section we present algorithms to compute MFG equilibria when the model is fully known.

\subsection{Subroutines}

The distribution computation for a given policy $\pi$ is described in Alg.~\ref{Algo:distrib-update} using forward (in time) iterations. The evaluation of the state-action value function for a \emph{given policy} $\pi$ and mean field flow $\mu$ is described in Alg.~\ref{Algo:valuefct-eval}.  The computation of the optimal value function  $Q^{*,\mu}$ for a given $\mu$ is described in Alg.~\ref{Algo:valuefct-opt}. A best response against $\mu$ can be obtained by running this algorithm and then taking an optimizer of $Q^{*,\mu}_n(x,\cdot)$ for each $n,x$.

\begin{algorithm}[tb]
   \caption{Forward update for the distribution}
   \label{Algo:distrib-update}
\begin{algorithmic}[1]
  \STATE {\bfseries Parameter:} policy $\pi = (\pi_n)_{n=0,\dots,N_T}$
  \STATE {\bfseries Output:} mean field flow $\mu = (\mu_n)_{n=0,\dots,N_T} = \mu^\pi = (\mu^\pi_n)_{n=0,\dots,N_T}$
   \STATE Let $\mu_0 = m_0$
   \FOR{$n=1\dots,N_T$}
   \STATE Let $\mu^{\pi}_{n} = P^{\mu^{\pi}_{n-1},\pi_{n-1}}_{n-1} \mu^{\pi}_{n-1}$
   \ENDFOR
   \STATE Return $\mu = (\mu_n)_{n=0,\dots,N_T}$
\end{algorithmic}
\end{algorithm}

\begin{algorithm}[tb]
   \caption{Backward induction for the value function evaluation}
   \label{Algo:valuefct-eval}
\begin{algorithmic}[1]
   \STATE {\bfseries Parameters:} policy $\pi = (\pi_n)_{n=0,\dots,N_T}$, mean field flow $\mu = (\mu_n)_{n=0,\dots,N_T}$
   \STATE {\bfseries Output:} state-action value function $Q^{\pi,\mu}$
   \STATE Let $Q_{N_T}(x,a) = r_{N_T}(x,a,\mu_{N_T})$
   \FOR{$n=N_T-1,\dots,0$}
   \STATE Compute 
   $$
    Q_{n}(x,a) = r_n(x,a,\mu_{n}) + \mathbb{E}_{x' \sim p_n(\cdot|x,a,\mu_n), a' \sim \pi_n(\cdot|x')}[Q_{n+1}(x',a')]
    $$ 
    where the expectation is computed in an exact way using the knowledge of the transition: 
            $$
                \mathbb{E}[Q_{n+1}(x',a')] = \sum_{x'} p_n(x'|x,a,\mu_n) \sum_{a'} \pi_{n+1}(a'|x') Q_{n+1}(x',a')
            $$
   \ENDFOR
   \STATE Return $Q = (Q_n)_{n=0,\dots,N_T}$
\end{algorithmic}
\end{algorithm}

\begin{algorithm}[tb]
   \caption{Backward induction for the optimal value function}
   \label{Algo:valuefct-opt}
\begin{algorithmic}[1]
   \STATE {\bfseries Parameters:} mean field flow $\mu = (\mu_n)_{n=0,\dots,N_T}$
   \STATE {\bfseries Output:} optimal state-action value function $Q^{*,\mu}$
   \STATE Let $Q_{N_T}(x,a) = r_{N_T}(x,a,\mu_{N_T})$
   \FOR{$n=N_T-1,\dots,0$}
   \STATE Compute 
   $$
    Q_{n}(x,a) = r_n(x,a,\mu_{n}) + \underbrace{ \mathbb{E}_{x' \sim p_n(\cdot|x,a,\mu_n)}[\max_{a'\in\actions} Q_{n+1}(x',a')] }_{=\sum_{x'} p_n(x'|x,a,\mu_n) \max_{a'} Q_{n+1}(x',a')}
    $$ 
    where the expectation is computed in an exact way using the knowledge of the transition $p_n$
   \ENDFOR
   \STATE {\bfseries Output:}  $Q = (Q_n)_{n=0,\dots,N_T}$
\end{algorithmic}
\end{algorithm}

\subsection{Main algorithms with fully known model}

Banach-Picard Fixed point iterations are presented in Alg.~\ref{alg:fixed-point}. Fictitious Play is described in Alg.~\ref{alg:Fictitious-play}. Policy iteration (for MFG) is presented in Alg.~\ref{alg:policy-iteration}. Online Mirror Descent is described in Alg.~\ref{Algo:OMD-exact}.

\begin{algorithm}[ht!]
\caption{Banach-Picard (BP) fixed point \label{alg:fixed-point}}
\begin{algorithmic}[1]
   \STATE {\bfseries Input:} Number of iterations $K$; optional softmax temperature $\eta$
   \STATE Initialize $\pi^0$
   \FOR{$k=0,\dots,K$}
      \STATE Forward update: Compute $\mu^k = \mu^{\pi^{k-1}}$, \textit{e.g.} using Alg.~\ref{Algo:distrib-update} with $\pi = \pi^{k-1}$\;
      \STATE Best response computation: Compute a BR $\pi^k$ against $\mu^k$, \textit{e.g.} by computing $Q^{*,\mu^k}$ using Alg.~\ref{Algo:valuefct-opt} with $\mu = \mu^k$  and then taking $\pi^k_n(\cdot|x)$ as a(ny) distribution over $\argmax Q^{*,\bar\mu^k}_n(x,\cdot)$ for every $n,x$; alternatively, compute a soft version: $\pi^k_n(\cdot|x) = \softmax(Q^{*,\mu^k}_n(x,\cdot)/\eta)$\;
    \ENDFOR
\STATE {\bfseries Output:} $\mu^K = (\mu^K_{n})_{n=0,\dots,N_T}$ and policy $\pi^K = (\pi^K_{n})_{n=0,\dots,N_T}$
\end{algorithmic}
\end{algorithm}

\begin{algorithm}[ht!]
\caption{Fictitious Play (FP) \label{alg:Fictitious-play}}
\begin{algorithmic}[1]
   \STATE {\bfseries Input:} Number of iterations $K$
   \STATE Initialize $\pi^0$
   \FOR{$k=0,\dots,K$}
      \STATE Forward update: Compute $\mu^k = \mu^{\pi^{k-1}}$, \textit{e.g.} using Alg.~\ref{Algo:distrib-update} with $\pi = \pi^{k-1}$\;
      \STATE Average distribution update: Compute $\bar \mu^k$ as the average of $(\mu^0, \dots, \mu^{\pi^{k}})$:  
      $$
        \bar \mu^k_n(x) = \tfrac{1}{k} \sum_{i=1}^k \mu^i_n(x) = \tfrac{k-1}{k}\bar\mu^{k-1}_n(x) + \tfrac{1}{k} \mu^k_n(x)
      $$
      \STATE Best response computation: Compute a BR $\pi^k$ against $\bar \mu^k$, \textit{e.g.} by computing $Q^{*,\bar\mu^k}$ using Alg.~\ref{Algo:valuefct-opt} and then taking $\pi^k_n(\cdot|x)$ as a(ny) distribution over $\argmax Q^{*,\bar\mu^k}_n(x,\cdot)$ for every $n,x$\; 
    \ENDFOR
\STATE {\bfseries Output:} $\bar \mu^K = (\bar\mu^K_{n})_{n=0,\dots,N_T}$ and policy $\bar \pi^K = (\bar\pi^K_{n})_{n=0,\dots,N_T}$ generating this mean field flow
\end{algorithmic}
\end{algorithm}

\begin{algorithm}[tb]
   \caption{Policy Iteration (PI)}
   \label{alg:policy-iteration}
\begin{algorithmic}[1]
   \STATE {\bfseries Parameters:} softmax temperature $\eta$; number of iterations $K$ 
   \STATE Initialize the sequence of tables $(\regq^0_{n})_{n=0,\dots,N_T}$, \textit{e.g.} with $\regq^0_{n}(x,a)=0$ for all $n,x,a$
   \STATE Let the projected policy be: $\pi^0_{n}(a|x) 
   = \softmax(\regq^0_{n}(x,\cdot)/\eta)(a)$ for all $n,x,a$
   \FOR{$k=1,\dots,K$}
       \STATE Forward Update: Compute $\mu^k = \mu^{\pi^{k-1}}$, \textit{e.g.} using Alg.~\ref{Algo:distrib-update} with $\pi = \pi^{k-1}$ 
       \STATE Backward Update: Compute $Q^k = Q^{\pi^{k-1}, \mu^{k}}$, \textit{e.g.} using backward induction Alg.~\ref{Algo:valuefct-eval} with $\pi = \pi^{k-1}$ and $\mu = \mu^k$ and then let $\pi^k_n(\cdot|x)$ be a(ny) distribution over $\argmax Q^{*,\bar\mu^k}_n(x,\cdot)$ for every $n,x$; alternatively, compute a soft version: $\pi^k_n(\cdot|x) = \softmax(Q^{k}_n(x,\cdot)/\eta)$\;
   \ENDFOR
   \STATE {\bfseries Output:} $Q^{K}, \pi^{K}$
\end{algorithmic}
\end{algorithm}

\begin{algorithm}[tb]
   \caption{Online Mirror Descent (OMD)}
   \label{Algo:OMD-exact}
\begin{algorithmic}[1]
   \STATE {\bfseries Parameters:} inverse learning rate parameter $\tau$; number of iterations $K$ 
   \STATE Initialize the sequence of tables $(\regq^0_{n})_{n=0,\dots,N_T}$, \textit{e.g.} with $\regq^0_{n}(x,a)=0$ for all $n,x,a$
   \STATE Let the projected policy be: $\pi^0_{n}(a|x) 
   = \softmax(\regq^0_{n}(x,\cdot))(a)$ for all $n,x,a$
   \FOR{$k=1,\dots,K$}
       \STATE Forward Update: Compute $\mu^k = \mu^{\pi^{k-1}}$, \textit{e.g.} using Alg.~\ref{Algo:distrib-update} with $\pi = \pi^{k-1}$
       \STATE Backward Update: Compute $Q^k = Q^{\pi^{k-1}, \mu^{k}}$, \textit{e.g.} using backward induction Alg.~\ref{Algo:valuefct-eval} with $\pi = \pi^{k-1}$ and $\mu = \mu^k$
          \STATE {Update the regularized $Q$-function and the projected policy: for all $n, x, a$,
       \begin{align*}
       \regq_{n}^{k}(x,a) 
       &= \regq_{n}^{k-1}(x,a) + \frac{1}{\tau} Q^{k}_n(x, a)
       \\
       \pi^{k}_{n}(a|x) 
       &= \softmax(\regq_{n}^{k}(x,\cdot))(a)
       \end{align*}
       }
   \ENDFOR
   \STATE Return $\regq^{K}, \pi^{K}$
\end{algorithmic}
\end{algorithm}

\clearpage

\section{Deep RL Algorithms}
\label{sec:deep-rl-algos}

We now present details on the deep RL algorithms used or developed in this paper. In this work, we focus on the use of deep RL for policy computation from the point of view of a representative agent. We assume that this agent has access to an oracle that can return $r_n(x_n,a_n,\mu_n)$ and a sample of $p_n(\cdot|x_n,a_n,\mu_n)$ when the agent follows is in state $x_n$ and uses action $a_n$. In fact, the collection of samples is split into episodes. At each episode, the agent start from some $x_0$ sampled from $m_0$. Then it evolves by following the current policy, and the transitions are added to a replay buffer.

In order to focus on the errors due to the deep RL algorithm, we assume that the distribution is updated in an exact way following Alg.~\ref{Algo:distrib-update}.

For the Deep RL part, the approaches can be summarized as follows:
\begin{itemize}
    \item Alg.~\ref{Algo:valuefct-eval}: Intuitively, he updates are replaced by stochastic gradient steps so as to minimize the following loss with respect to $\theta$:
    \begin{equation}
        \label{eq:loss-eval-Q}
        \left|Q_{\theta}((n,x_n),a_n) - r_n(x_n,a_n,\mu_{n}) - \hat{\mathbb{E}}_{x'_{n+1} \sim p_n(\cdot|x_n,a_n,\mu_n), a'_{n+1} \sim \pi_{n+1}(\cdot|x'_{n+1})}[q((n+1,x'_{n+1}),a'_{n+1})]\right|^2,
    \end{equation}
    where $\hat{\mathbb{E}}$ denotes an empirical expectation over a finite number of samples picked from the replay buffer and $q$ is replaced by a target network $Q_{\theta'}$ whose parameters are frozen while training $\theta$ and that are updated less frequently than $\theta$.  Combined with Policy Iteration (Alg.~\ref{alg:policy-iteration}), this leads to the algorithm referred to as \textbf{Deep Policy Iteration (D-PI)}.
    \item Alg.~\ref{Algo:valuefct-opt}: We can proceed similarly, except that the target becomes:
    $$
        r_n(x_n,a_n,\mu_{n}) + \hat{\mathbb{E}}_{x'_{n+1} \sim p_n(\cdot|x_n,a_n,\mu_n)}[\max_{a'}q((n+1,x'_{n+1}),a')].
    $$
    In fact, in our implementation we use DQN~\citep{mnih2013playingatari} as a subroutine for the BR computation. Combined with Banach-Picard iterations (Alg.~\ref{alg:fixed-point}), this leads directly to the algorithm referred to as \textbf{Deep Banach-Picard (D-BP)}.
    \item To obtain \textbf{Deep Average-network Fictitious Play (D-AFP)} (Algo.~\ref{algo:deep-FP-avgBR}), at each iteration, the best response against the current distribution is learnt using DQN~\citep{mnih2013playingatari}. This policy is used to generate trajectories, whose state-action samples are added to a reservoir buffer $\mathcal{M}_{SL}$. This buffer stores state-actions generated using past policies from previous iterations.  Then, an auxiliary neural network for the logits representing the average policy is trained using supervised learning using $\mathcal{M}_{SL}$: stochastic gradient is used to find $\bar\theta$  minimizing approximately the loss:
      \begin{equation*}
        \mathcal{L}(\bar\theta) = \EE_{(s,a) \sim \mathcal{M}_{SL}}\left[-\log \left( \bar\pi_{\bar\theta}(a|s)\right)\right]
      \end{equation*}
      In our implementation, for the representation of the average policy, we use a neural network $\bar \ell_{\omega}$ with parameters $\omega$ for the logits, and then we compute the policy as: $\bar\pi = \softmax(\bar \ell_{\omega})$. 
        This step is reminiscent of Neural Fictitious Self Play (NFSP) introduced in~\citet{heinrichsilver2016deepnfsp}, which was used to solve Leduc poker and Limit Texas Hold’em poker. 
        However, the overall algorithm is different. Indeed, in NFSP as described in Algorithm~1 of~\citet{heinrichsilver2016deepnfsp}, the neural network for the average policy and the neural network for the Q-function are both updated at each iteration. We reuse the idea of having a buffer of past actions but in our case, between each update of the average policy network, we do two operations: first, we update the mean field (sequence of population distributions) and second, we learn a best response against this mean field.
    \item To obtain \textbf{Deep Munchausen Online Mirror Descent (D-MOMD)} (Algo.~\ref{Algo:M-OMD-exact}), we can simply modify the target in~\eqref{eq:loss-eval-Q} as follows:
   \begin{align*} 
        \Big|Q_{\theta}((n,x_n),a_n) &- r_n(x_n,a_n,\mu_{n}) \, {\color{red} -\tau\ln\pi_{n-1}(a_n|x_n) }
        \\
        &- \hat{\mathbb{E}}_{x'_{n+1} \sim p_n(\cdot|x_n,a_n,\mu_n), }\sum_{a'}\pi_n(a'|x_n)[q((n+1,x'_{n+1}),a'_{n+1}) \,{\color{blue}- \tau \ln\pi_n(a'|x'_{n+1})}]\Big|^2,
    \end{align*} 
    where $q = Q_{\theta'}$ is a target network whose parameters $\theta'$ are frozen while training $\theta$. This is similar to equation (7) of~\citet{vieillard_munchausen_neurips_2020} which introduced the Munchausen RL method. However, in our case the distribution $\mu_n$ appears in the reward $r_n$ and in the transition leading to the new state $x'_{n+1}$. So while \citet{vieillard_munchausen_neurips_2020} simply repeatedly update the Q-network, we intertwine the udpates of the cumulative Q-network with the updates of the population distribution.
\end{itemize}

\clearpage
\section{Details on the link between MOMD and regularized MDPs}
\label{sec:details-Munchausen-alpha}

Consider regularizing the MFG with only entropy, that is
\begin{equation}
    J(\pi,\mu) = \EE_{\pi}\left[\sum_{n=0}^{N_T}(r_n(s_n,a_n,\mu_n) - (1-\alpha)\tau \ln \pi_n(a_n|s_n))\right].
\end{equation}
Notice that we choose $(1-\alpha)\tau$ here because it will simplify later, but it would work with any temperature (or learning rate from the OMD perspective).

Now, let's solve this MFG with OMD with learning rate $(\alpha\tau)^{-1}$, adopting the KL perspective. The corresponding algorithm is:
\begin{align}
    &\pi_{n}^{k+1} \in \argmax \langle \pi_n, q_{n}^{k}\rangle - \alpha\tau \mathrm{KL}(\pi_n || \pi_{n}^{k}) + (1-\alpha)\tau \mathcal{H}(\pi_n)
    \\
    &\begin{cases}
        q_{N_T}^{k+1} = r_{N_T}^{k+1}
        \\
        q_{n}^{k+1} = r_{n}^{k+1} + \gamma P \langle \pi_{n+1}^{k+1}, q_{n+1}^{k+1} - (1-\alpha)\tau \ln \pi_{n+1}^{k+1}\rangle 
    \end{cases}
\end{align}
Next, using $Q_{n}^{k} = q_{n}^{k} + \alpha \tau \ln \pi_{n}^{k}$, we can rewrite the evaluation part as:
\begin{align}
    &\pi_{n}^{k+1} = \softmax\left(\frac{Q_{n}^{k}}{\tau}\right)
    \\
    &\begin{cases}
        Q_{N_T}^{k+1} = r_{N_T}^{k+1} + \alpha \tau \ln \pi_{N_T}^{k+1}
        \\
        Q_{n}^{k+1} = r_{n}^{k+1} + \alpha\tau\ln\pi_{n}^{k+1} + \gamma P \langle \pi_{n+1}^{k+1}, Q_{n+1}^{k+1} - \tau \ln \pi_{n+1}^{k+1}\rangle 
    \end{cases}
\end{align}
We remark that it corresponds to the ``scaled'' version of Munchausen OMD, meaning that it amounts to solving the MFG regularized with $(1-\alpha)\tau\mathcal{H}(\pi)$ with OMD. We retrieve with $\alpha=1$ the unscaled version of Munchausen OMD, addressing the unregularized MFG. It also makes a connection with the Boltzmannn iteration method of \citet{pmlr-cui21-approximately}, in which a similar penalization except that the penalty involves a fixed policy instead of using the current policy. Their prior descent method, in which the reference policy is updated from time to time, can thus be viewed as a first step towards Munchausen OMD.

\clearpage
\section{Hyperparameters sweeps}
\label{sec:hyperparams-sweeps}

\begin{figure}[tbh]
    \centering
    \begin{minipage}{0.9\linewidth}
    \includegraphics[width=.85\linewidth]{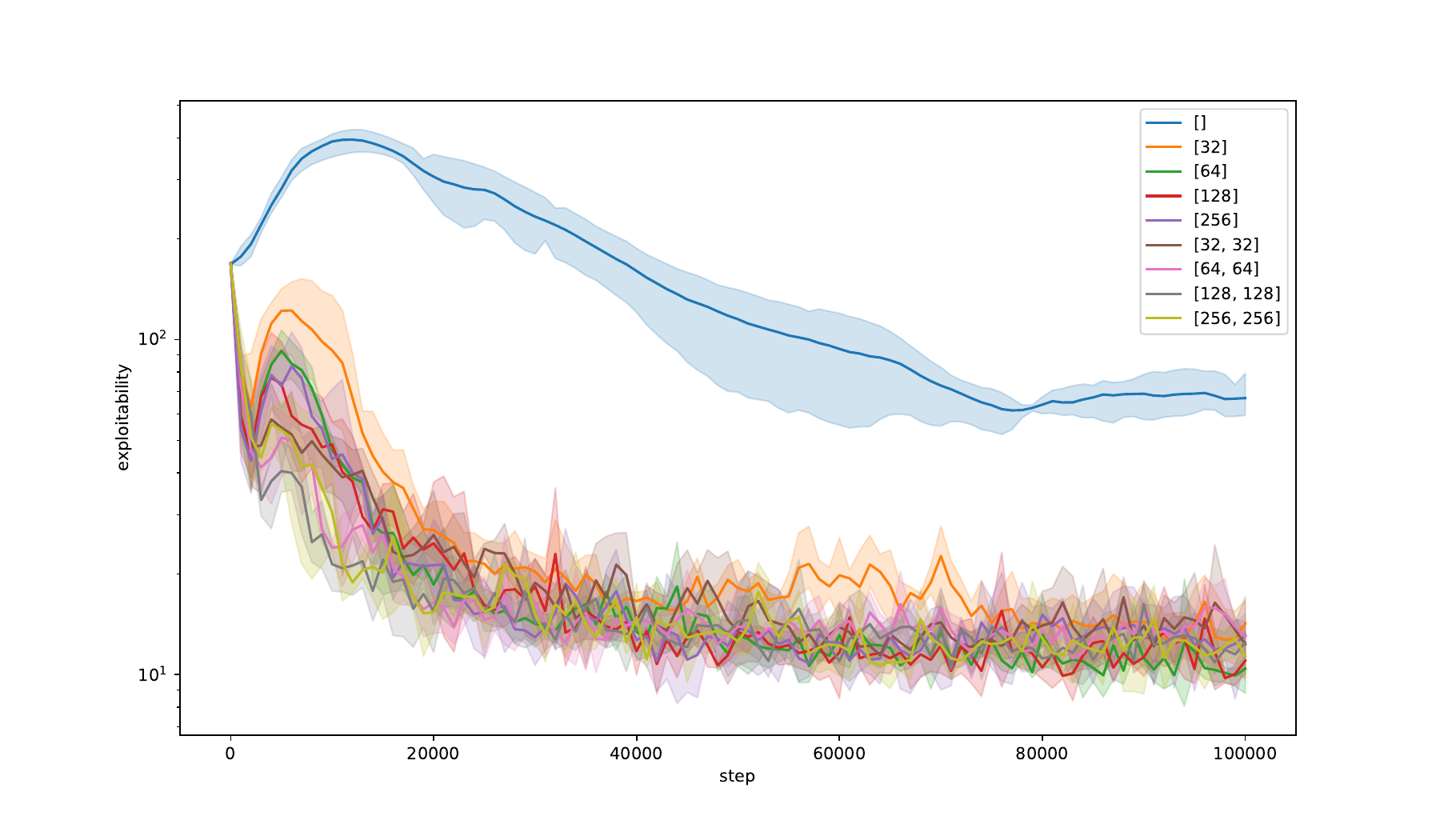} 
    \end{minipage}
    \caption{
    D-MOMD, Exploration game with four rooms: Sweep over the network size. The neural network architecture is feedforward fully connected with one or two hidden layers, except for the curve with label $\texttt{[]}$, which refers to a linear function. This illustrates in particular that the policy can not be well approximated using only linear functions, hence the need for non-linear approximations, which raises the difficulty of averaging or summing such approximations (here neural networks).
    }
    \label{fig:curl_example_sweep_network}
\end{figure}

\begin{center}
\begin{figure}[tbh]
    \centering
    \begin{minipage}{1\linewidth}
    \includegraphics[width=.85\linewidth]{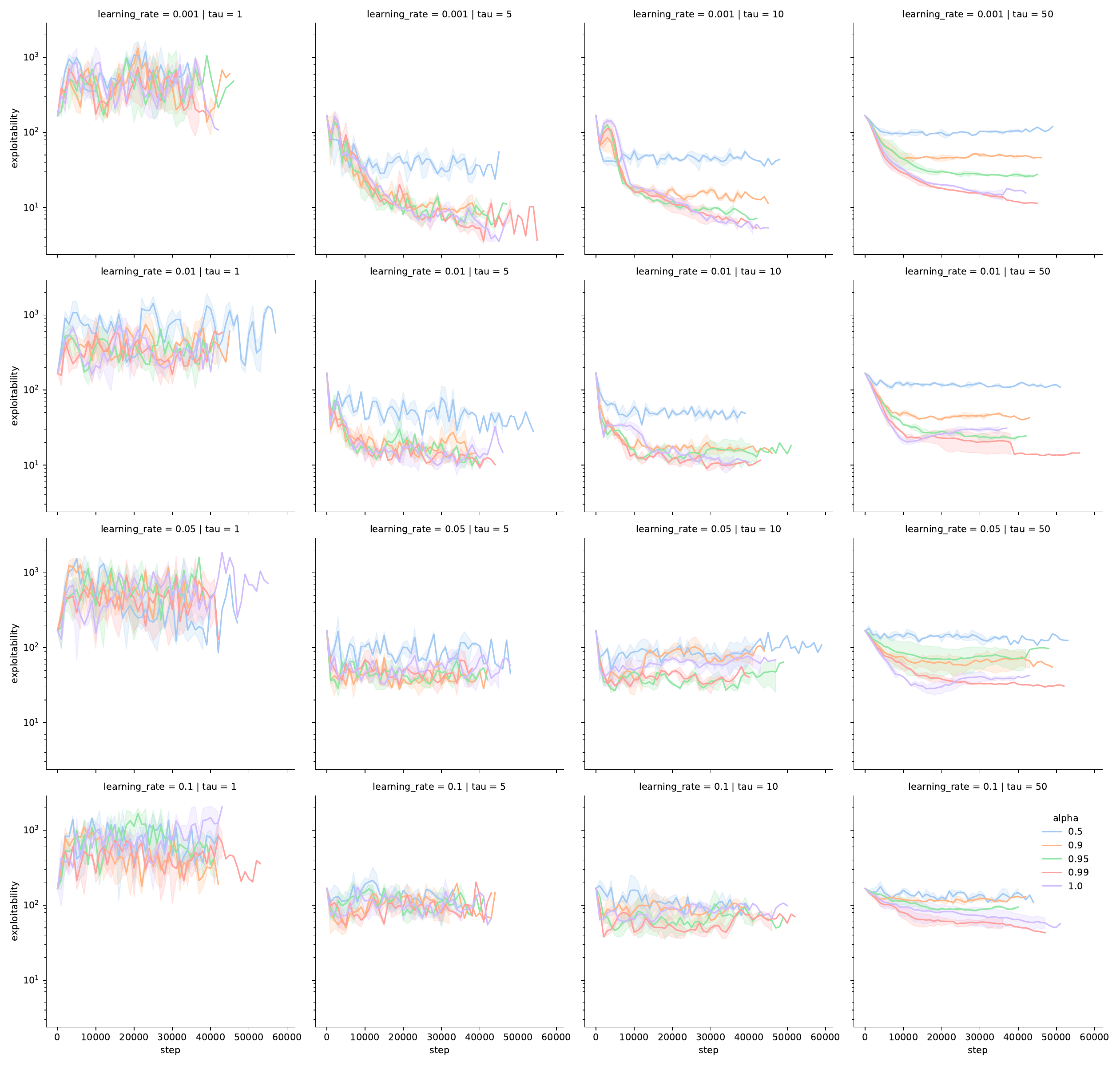} 
    \end{minipage}
    \caption{ 
    D-MOMD, Exploration game with four rooms: Sweep over $\tau, \alpha$ and learning rate.  
    }
    \label{fig:curl_example_sweep_tau_alpha_lr}
\end{figure}
\end{center}

\end{document}